%% file: main.tex
\documentclass[letterpaper, 10 pt, conference]{ieeeconf}
\usepackage{times}
\usepackage{graphicx}
\usepackage{amsmath,amssymb,amsopn,amstext,amsfonts}
\usepackage{cancel}
\usepackage[space]{cite}
\usepackage{pdfsync}
\usepackage{balance}
\usepackage{color}
\usepackage{mathtools}
\usepackage{stfloats}
\usepackage{algpseudocode}
\usepackage{xspace}
\usepackage{algorithm}
\usepackage{bm}

\usepackage{diagbox}
\usepackage{float}
\usepackage{epstopdf}
\usepackage{pifont}
\usepackage{multirow}
\usepackage{url}
\usepackage{verbatim}
\makeatletter
\let\NAT@parse\undefined
\makeatother
\usepackage[linkcolor=black,citecolor=black,urlcolor=black,colorlinks=TRUE]{hyperref}
\usepackage{booktabs}
\usepackage{subcaption}
\usepackage{caption}
\usepackage{cuted}

\usepackage{makecell}
\usepackage{array}

\input{preamble}

\graphicspath{{./}}
\DeclareGraphicsExtensions{.png,.jpg,.eps,.pdf}
\IEEEoverridecommandlockouts
\title{\LARGE \bf \methodname: Asynchronous-to-Global 3D Reconstruction from Event Camera via Spatial-Temporal Feature Aggregation}
\author{Jian Huang$^{1,2 *}$, Haotian Shen$^{2 *}$, Xinhao Lou$^{2 *}$, Chengrui Dong$^{1,2}$, Wenpu Li$^{2}$, Peidong Liu$^{2 \dagger}$
    \thanks{$^{*}$Equal Contribution.}
    \thanks{$^{\dagger}$Corresponding author.}
    \thanks{$^{1}$Jian Huang and Chengrui Dong are with Zhejiang University, Hangzhou, Zhejiang, China.}
    \thanks{$^{2}$All authors are with Westlake University, Hangzhou, Zhejiang, China. E-mails: {\ttfamily\small \{huangjian39, shenhaotian, louxinhao, dongchengrui, liwenpu, liupeidong\}westlake.edu.cn}}
    \vspace{-1.8em}
}

\begin{document}

\IEEEaftertitletext{%
    \vspace{-2.0em}
    \begin{center}
        \includegraphics[width=0.80\textwidth]{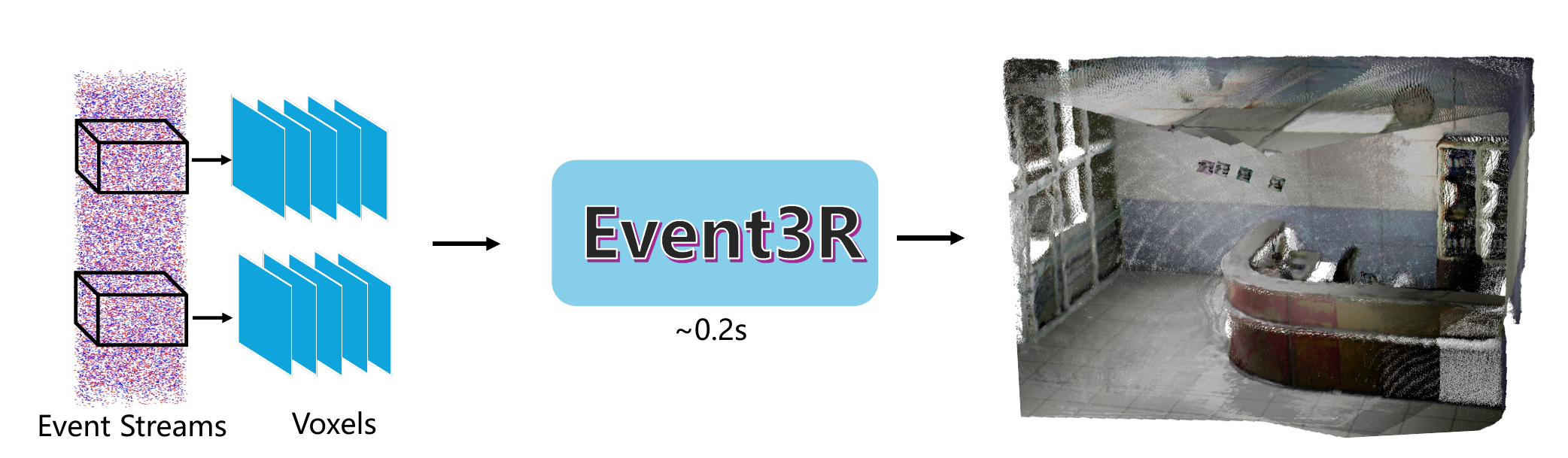}
        \vspace{0.3em}
        {\footnotesize
        \captionof{figure}{\textbf{Illustration of our \methodname.} \methodname takes two or more short segments of asynchronous event streams, converts them into spatial–temporal voxel representations, and directly predicts a globally aligned 3D pointmap in about 0.2s. For visualization, the pointcloud is colored using the RGB image corresponding to the center voxel bin, although no RGB data is used during training or inference.}}
    \end{center}
    \vspace{0.5em}
}
\maketitle

\begin{abstract}

	Robust 3D reconstruction is essential for robotics and embodied perception. Recent feed-forward approaches such as DUSt3R have demonstrated impressive progress in dense 3D reconstruction from RGB images, achieving global geometric consistency and strong generalization. However, extending such dense 3D reconstruction to event cameras remains challenging due to their asynchronous, sparse, and highly dynamic nature, as well as the lack of large-scale, well-labeled datasets.
	In this work, we introduce \methodname, a feed-forward framework that directly maps asynchronous event streams to globally consistent 3D point clouds. Event3R represents incoming events as spatial–temporal voxels, enabling time-aware feature integration through a temporal attention module that enhances module's temporal feature learning.
	To further strengthen temporal representation learning and reduce reliance on labeled data, we propose a Masked Bin Modeling (MBM) strategy for self-supervised pretraining, enabling robust temporal representation learning with minimal labeled data, and retain it as an auxiliary fine-tuning objective. 
	In addition, contrastive alignment and consistency regularization losses are incorporated during fine-tuning to reinforce structural correspondence and temporal coherence across views.
	Extensive experiments on both synthetic and real-world benchmarks demonstrate that Event3R achieves robust, temporally consistent, and globally aligned 3D reconstructions, significantly outperforming existing event-based methods.

\end{abstract}

\IEEEpeerreviewmaketitle
\section{Introduction}
\label{sec:introduction}

\begin{figure}
    \centering
    \setlength{\tabcolsep}{1pt}
    \footnotesize
    \setlength{\tabcolsep}{1pt}
	\begin{tabular}{p{8.2pt}cccc}
		&{Event Input} & {RGB Input} & {Event3R} & {DUSt3R}  \\
		\rotatebox[origin=c]{90}{\footnotesize {HDR}} & 		
        \parbox[c]{0.23\linewidth}{\vspace{1pt}\centering\includegraphics[width=\linewidth]{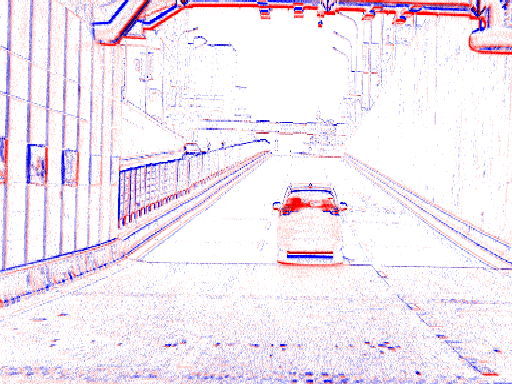}} &
		\parbox[c]{0.23\linewidth}{\vspace{1pt}\centering\includegraphics[width=\linewidth]{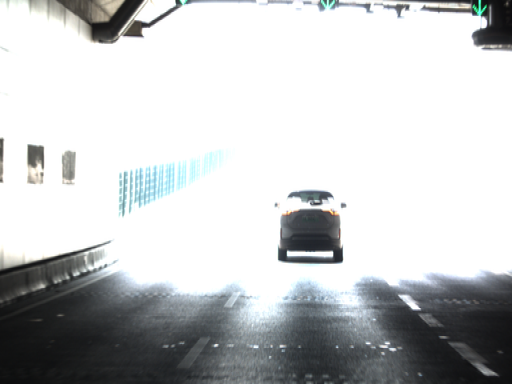}} &
		\parbox[c]{0.23\linewidth}{\vspace{1pt}\centering\includegraphics[width=\linewidth]{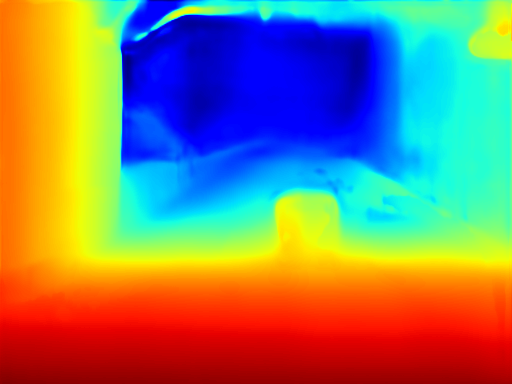}} &
		\parbox[c]{0.23\linewidth}{\vspace{1pt}\centering\includegraphics[width=\linewidth]{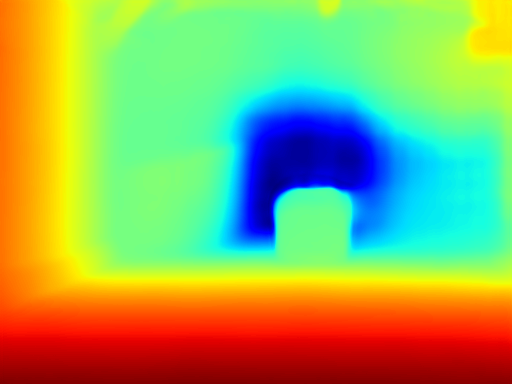}}

        \\
		\rotatebox[origin=c]{90}{\footnotesize {Low Light}} & 
        \parbox[c]{0.23\linewidth}{\vspace{1pt}\centering\includegraphics[width=\linewidth]{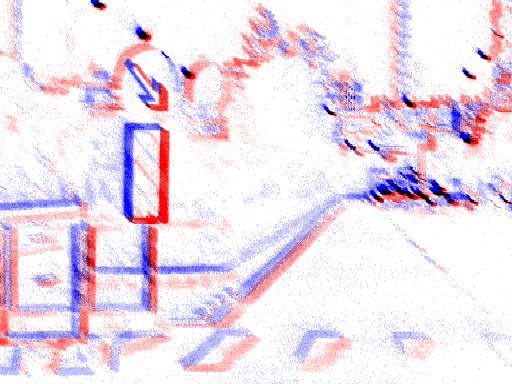}} &
		\parbox[c]{0.23\linewidth}{\vspace{1pt}\centering\includegraphics[width=\linewidth]{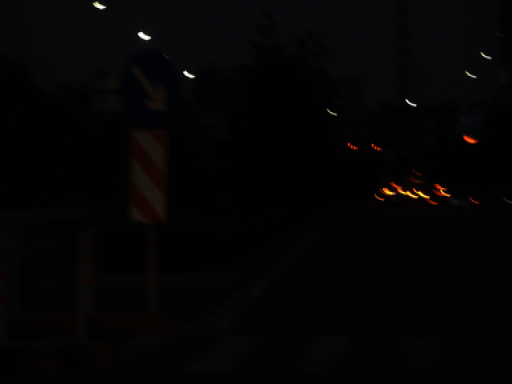}} &
		\parbox[c]{0.23\linewidth}{\vspace{1pt}\centering\includegraphics[width=\linewidth]{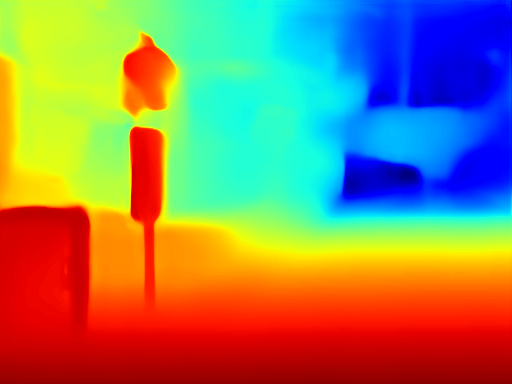}} &
		\parbox[c]{0.23\linewidth}{\vspace{1pt}\centering\includegraphics[width=\linewidth]{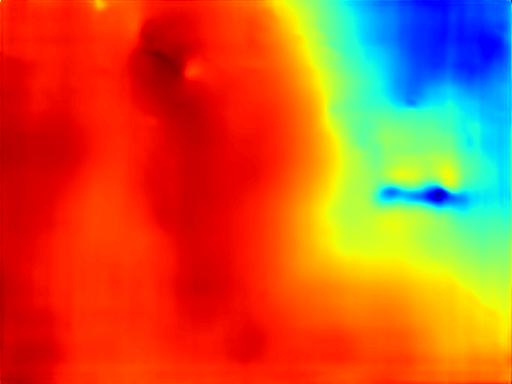}}

	\end{tabular}
    \vspace{-3pt}
    \caption{\textbf{Qualitative depth estimation comparison with DUSt3R under challenging conditions}. Event3R is evaluated in a zero-shot setting without training on the PEOD \cite{cui2025peod} dataset.}
    \label{fig:vsDust3R}
    \vspace{-3.5em}
\end{figure}

Robust 3D reconstruction is a fundamental task for robotics, AR/VR, and autonomous systems, providing dense geometric understanding for navigation, interaction, and scene interpretation. Traditional optimization-based approaches such as structure-from-motion (SfM) and multi-view stereo (MVS) recover accurate geometry but require iterative optimization and are prone to failure in dynamic or textureless environments. NeRF \cite{mildenhall2021nerf} and 3D-GS \cite{kerbl3Dgaussians} based methods achieve photorealistic quality, yet their per-scene optimization severely limits efficiency and generalization.
Recently, feed-forward 3D reconstruction frameworks such as DUSt3R \cite{wang2024dust3r}, CUT3R\cite{wang2025continuous} and VGGT \cite{wang2025vggt} have demonstrated remarkable progress by predicting globally consistent 3D structures directly from RGB frames, offering orders-of-magnitude faster inference without iterative refinement. However, their reliance on frame-based images makes them vulnerable under fast motion or challenging lighting conditions, where conventional cameras fail to capture sufficient information. This limitation is illustrated in Fig.~\ref{fig:vsDust3R}, where DUSt3R struggles under challenging illumination due to degraded RGB observations, while Event3R remains robust.

Event cameras offer a promising alternative sensing modality. Unlike conventional cameras that record absolute intensity frames at fixed rates, event cameras asynchronously capture per-pixel brightness changes with microsecond temporal precision, enabling high dynamic range, low latency, and motion blur–free perception. These properties make them ideal for challenging scenarios, as demonstrated by recent event-based 3D reconstruction methods such as EvGGS\cite{wang2024evggs}, Event3DGS\cite{xiong2024event3dgs} and IncEventGS\cite{huang2025inceventgs}.
Yet, most of these methods focus on local depth(GS) prediction or rely on optimization-based rendering, lacking a global, feed-forward event-based reconstruction approach. However, extending DUSt3R-style reasoning to asynchronous, sparse event streams remains an open challenge due to their unique spatiotemporal structure and the scarcity of large well-labeled event–depth–pose datasets.

To fill this research gap, we present \methodname, a feed-forward framework designed to bring global 3D reconstruction to asynchronous event data. Our key insight is to reinterpret event streams as structured spatial–temporal voxels, allowing geometric reconstruction backbone \ie DUSt3R to process temporally dense yet spatially sparse observations. Building on this representation, a temporal attention mechanism models motion information across voxelized time bins, enabling globally consistent point cloud reconstruction.
To further enhance temporal representation learning with limited labeled data, we introduce Masked Bin Modeling (MBM) — a self-supervised pretraining objective that regularizes fine-tuning and improves motion generalization. Combined with contrastive alignment and temporal consistency losses to enforce cross-view coherence, Event3R provides a new demonstration of globally consistent 3D reconstruction from purely event-based inputs, effectively bridging asynchronous sensing and feed-forward 3D reconstruction.

In summary, our main contributions are as follows:

\begin{itemize}
    \item We present \methodname, a new feed-forward framework that enables instant global 3D reconstruction from asynchronous event streams.
    \item We introduce a temporal attention module that enables time-aware feature integration across voxelized time bins.
    \item We propose Masked Bin Modeling (MBM), a self-supervised pretraining strategy that improves temporal information aggregation and reduces the reliance on large amounts of labeled data.
    \item We validate \methodname on multiple challenging 3D perception tasks, including dense reconstruction, depth estimation, and pose estimation, demonstrating significant improvements in downstream performance and practical impact.
\end{itemize}

\section{Related Works}
\label{sec:relatedwork}

We review two main areas of prior work: RGB-based and event-based 3D reconstruction, which represent the most relevant components for our study.

\PAR{RGB-based 3D Reconstruction.}
Dense 3D reconstruction from RGB imagery has a long history, ranging from classical optimization pipelines (structure-from-motion and multi-view stereo) to modern learning-based systems. Conventional toolchains \cite{colmap, xun2023crepes} recover accurate geometry by sparse feature matching and bundle adjustment, but remain computationally intensive and fragile in dynamic or textureless scenes. Neural scene representations broadened the design space: neural radiance fields (NeRF) and follow-ups (e.g., mip-NeRF) deliver photorealistic novel-view synthesis via per-scene optimization \cite{mildenhall2021nerf, barron2021mip}, while explicit point/feature-based schemes such as 3D Gaussian Splatting (3D-GS) accelerate rendering and enable real-time performance for high-fidelity reconstructions \cite{kerbl3Dgaussians}. While 3D-GS achieves efficient and high-fidelity reconstruction from multi-view images with known camera poses, subsequent efforts have begun relaxing this requirement. 

COLMAP-Free 3D-GS \cite{fu2024colmap} sequentially grows a 3D Gaussian set directly from input frames without pose precomputation, simplifying the reconstruction pipeline and reducing preprocessing overhead.

More recently, a new feed-forward paradigm has emerged that internalizes the SfM+MVS pipeline into a single inference pass: seminal works like DUSt3R\cite{wang2024dust3r} and subsequent models (e.g., MASt3R\cite{leroy2024grounding},  CUT3R\cite{wang2025continuous}, VGGT\cite{wang2025vggt}) employ dense correspondence learning, transformer-based global reconstruction, and pose-from-alignment mechanisms to jointly infer camera poses and dense geometry from unconstrained image sets \cite{zhang2025review}. This family achieves orders-of-magnitude speedups and improved robustness in low-texture scenarios by leveraging learned priors and large training corpora, but it also shifts the bottleneck from iterative geometric solvers to GPU memory, model scaling, and dataset scale/diversity. 
Crucially, these feed-forward methods are designed for synchronous, frame-based inputs; their reliance on intensity frames limits performance in extreme motion or challenging lighting conditions and leaves open the problem of bringing feed-forward global 3D reconstruction to asynchronous event streams.

\PAR{Event-based 3D Reconstruction.} 
Early works on event-based 3D reconstruction primarily focused on probabilistic filtering and geometric mapping. Kim \etal~\cite{kim2016real} proposed a joint optimization framework using three decoupled probabilistic filters to estimate 6-DoF camera motion, scene intensity gradients, and inverse depth relative to a keyframe. Following this line, Rebecq \etal~\cite{rebecq2018emvs} introduced EMVS, a real-time event-based multi-view stereo pipeline that reconstructs semi-dense depth maps from asynchronous events. Subsequent semi-dense reconstruction and mapping systems such as EVO~\cite{Rebecq2017ral} and Stereo Event Reconstruction~\cite{zhou2018semi} further improved geometric accuracy and robustness by tightly coupling depth estimation with pose tracking. While these methods demonstrated the unique potential of event cameras for 3D perception, they remain fundamentally optimization-based, requiring iterative updates and handcrafted motion models that limit scalability and global consistency. 

To overcome these limitations, recent efforts have extended neural implicit representations to event-driven input. Klenk \etal~\cite{klenk2023nerf}, Hwang \etal~\cite{hwang2023ev}, and Rudnev \etal~\cite{rudnev2023eventnerf} concurrently proposed Event-NeRF variants that recover neural radiance fields from event streams given known camera trajectories. Building on this, Event-GS~\cite{xiong2024event3dgs, yura2025eventsplat, han2024event} adapted 3D Gaussian Splatting to asynchronous event data, while IncEventGS~\cite{huang2025inceventgs} further relaxed the dependence on external pose estimation through incremental optimization. Although these methods successfully transfer modern neural representations to event data, they still rely on explicit or estimated camera poses and per-scene optimization, which hinders efficiency and generalization. 

More recently, a trend has emerged toward pose-free and feed-forward event-based reconstruction. Feed-forward approaches such as EvGGS~\cite{wang2024evggs} first predict depth and local Gaussian splats from event input, then use ground-truth camera poses to transform these Gaussians into a common coordinate frame — a design that constrains global reconstruction and limits generalization across diverse scenes. Concurrently, EAG3R~\cite{wueag3r} augments RGB-based feed-forward pipelines with pretrained event encoders to enhance robustness under extreme conditions. However, it neither fully exploits the temporal dynamics intrinsic to event streams nor addresses the scarcity of large-scale labeled event datasets. These limitations highlight the need for a unified, feed-forward framework capable of producing globally consistent 3D reconstructions directly from asynchronous event data.

\section{Method}

Our method performs feed-forward 3D reconstruction directly from event streams. 
It takes two or multiple ($N>2$) asynchronous event sequences as input and produces a globally aligned 3D pointmap within one second, without relying on iterative optimization or external pose estimation. 
To achieve this, we propose a unified and efficient framework that bridges the gap between sparse, asynchronous event data and dense geometric reconstruction.

Specifically, we first convert the incoming events into structured spatial–temporal voxel tensors, which preserve fine-grained temporal dynamics while enabling parallel processing. 
A temporal attention encoder is then employed to model inter-bin dependencies and aggregate motion cues across time, capturing both local dynamics and long-range correlations. 
These temporally enriched features are further processed by a DUSt3R-based reconstruction backbone to infer dense depth and camera-aligned 3D pointmaps with strong geometric consistency.

\begin{figure*}[b]
	\centering
	\includegraphics[width=0.90\linewidth]{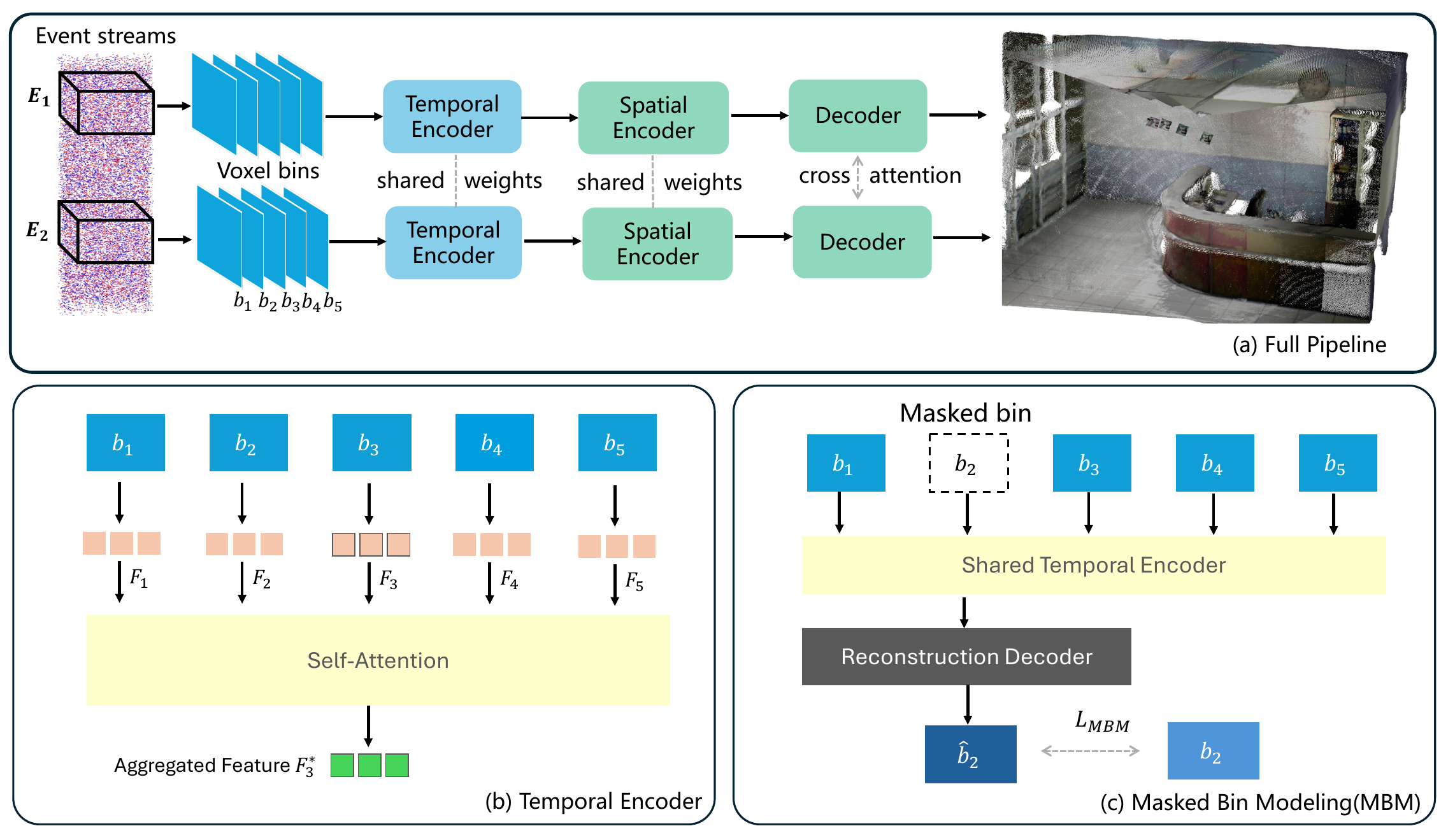}
	\caption{\textbf{Overview of the \methodname pipeline.} 
		(a) Full pipeline: asynchronous event streams are converted into spatial–temporal voxels (T bins) and processed by the temporal encoder to capture motion cues. The resulting features are then passed through a spatial encoder and decoder, built on a DUSt3R backbone, to predict dense, globally aligned 3D pointmaps. 
		(b) Temporal Encoder: self-attention across voxelized time bins enables temporal feature learning while preserving spatial alignment. 
		(c) To further enhance temporal representation, Masked Bin Modeling (MBM) is applied in two stages: first, self-supervised pretraining teaches the model to learn temporal cues and alleviates the scarcity of labeled event datasets; second, MBM serves as an auxiliary loss during joint training, reinforcing temporal feature learning.}
	\label{fig2}
\end{figure*}

In addition, we introduce a Masked Bin Modeling (MBM) strategy that promotes temporal feature learning through self-supervision, as well as contrastive and consistency objectives that enforce spatial coherence and improve generalization across diverse scenes. 
Together, these designs enable efficient, scalable, and label-efficient 3D reconstruction from event streams in a fully feed-forward manner. 
An overview of the proposed pipeline is shown in Fig.~\ref{fig2}, and the following sections detail each component.

\subsection{Event-to-Voxel Representation}
Each event $e_k = (u_k, t_k, p_k)$ is triggered asynchronously at pixel location $u_k = (x_k, y_k)$ with timestamp $t_k$ and polarity $p_k \in \{+1, -1\}$, indicating the sign of brightness change. 
To enable synchronous processing, we discretize the continuous event stream within a short temporal window $\Delta t$ into a voxel grid of size $H \times W \times T$, where $T$ denotes the number of temporal bins. 
Each event is then splatted into its neighboring bins using trilinear interpolation according to its normalized timestamp:
\begin{equation}
	E(x, y, n) = \sum_k p_k \, \max(0, 1 - |n - t_k^*|),
\end{equation}
where $t_k^* = \frac{T-1}{\Delta t} (t_k - t_0)$ represents the normalized temporal coordinate.
This voxelization yields a dense yet time-preserving spatial–temporal representation, effectively bridging the gap between the asynchronous event stream and feed-forward 3D reconstruction.

\subsection{Temporal Encoder(TE)}

To effectively exploit the temporal dynamics in event voxel data, we introduce a \textbf{Temporal Encoder} that aggregates information across multiple time bins while maintaining spatial alignment. Given an event voxel input $\mathbf{V} \in \mathbb{R}^{B \times T \times H \times W}$ with $T$ temporal bins, 
the encoder produces a temporally enriched feature map compatible with the DUSt3R backbone. 
We note that $T$ is set to an odd number to ensure a well-defined \textit{central} bin, which serves as the temporal reference frame for predicting the final pointmap. 
This design provides symmetry around the reference bin and enables balanced aggregation of past and future event information.

\textbf{Per-bin Feature Extraction.}
Each temporal bin $\mathbf{V}_t \in \mathbb{R}^{B \times 1 \times H \times W}$ is treated as an image-like frame that summarizes events within a short, fixed-duration time slice. 
Unlike conventional RGB frames, these bins are \textit{strictly ordered in time}, preserving the intrinsic temporal structure of the event stream. This ordering is crucial for temporal reasoning, as it allows the encoder to model motion dynamics and brightness changes by attending across consecutive bins in a temporally consistent manner.

Instead of collapsing each bin into a global feature, we decompose it into non-overlapping spatial \textit{patches}, preserving fine-grained geometric details while reducing the computational cost of temporal attention by representing each bin as a compact set of patch tokens. Formally, each temporal bin $\mathbf{V}_t$ is embedded via a convolutional layer:
\begin{equation}
	\mathbf{F}_t = \mathrm{ConvEmbed}(\mathbf{V}_t) \in \mathbb{R}^{B \times C_e \times H_p \times W_p}.
\end{equation}
Here, $C_e$ is the embedding dimension, and $H_p = H / p$ and $W_p = W / p$ denote the spatial dimensions of the patch grid, with $p$ being the patch size. 
The resulting patch-level feature map $\mathbf{F}_t$ maintains local spatial coherence while providing an efficient representation for temporal modeling.

\paragraph{Temporal Attention.}
To capture inter-bin temporal information, we first stack patch sequences from all $T$ bins along the temporal dimension:
\begin{equation}
	\mathbf{X} \in \mathbb{R}^{B \times N_p \times T \times C_e}, \quad N_p = H_p \cdot W_p.
\end{equation}
We then reshape $\mathbf{X}$ to $(B \cdot N_p, T, C_e)$ so that each spatial patch is processed independently by the temporal encoder.

For each patch position $i$, the encoder performs self-attention over the $T$ temporal bins. 
Denoting the temporal sequence for the $i$-th patch as
\begin{equation}
	\mathbf{X}_i = [\mathbf{F}_{i}^{1}, \dots, \mathbf{F}_{i}^{T}] \in \mathbb{R}^{T \times C_e},
\end{equation}
the aggregated feature of the center bin (index $t_\text{center} = (T+1)/2$) is computed as
\begin{equation}
	\mathbf{F}_{i, \text{center}}^* = \mathrm{Attention}(\mathbf{Q} = \mathbf{F}_{i}^{t_\text{center}}, \; \mathbf{K}, \mathbf{V} = \mathbf{X}_i),
\end{equation}
where $\mathbf{K}, \mathbf{V} = \mathbf{X}_i$ are the keys and values from all temporal bins. 
In this way, the center bin feature $\mathbf{F}_{i, \text{center}}^*$ aggregates information from its neighboring bins, effectively capturing inter-bin temporal dynamics. 
The temporal encoder consists of $L$ layers of multi-head self-attention and feed-forward networks.

These patch-level features, now enriched with temporal context, are directly passed to the spatial backbone for further processing, enabling efficient and compatible spatio-temporal representation learning.

\subsection{Masked Bin Modeling}

To further enhance temporal feature learning, we adopt a masked bin modeling (MBM) strategy. 
Given the temporally aggregated feature map 
\(
\mathbf{F}^* \in \mathbb{R}^{B \times C_e \times H \times W}
\)
from the temporal encoder, MBM randomly masks features from one or more bins and trains the network to reconstruct the masked content from the remaining visible bins. 
Different from prior masked event modeling \cite{klenk2024masked}, which targets event/token-level self-supervised pretraining, our MBM performs bin-level masking and is designed for geometry-aware representation learning. It explicitly enforces inter-bin temporal modeling crucial for depth and pose estimation.

Let \(M_t \in \{0,1\}^{B \times 1 \times H \times W}\) denote the binary mask applied to bin \(t\), and let \(\mathbf{F}^*_{\text{masked}} = \mathbf{F}^* \odot (1-M_t)\) be the masked feature map. 
A lightweight reconstruction decoder \(\mathcal{D}\) predicts the masked bin feature:
\begin{equation}
	\hat{\mathbf{F}}_t = \mathcal{D}(\mathbf{F}^*_{\text{masked}}),
\end{equation}
with the reconstruction loss
\begin{equation}
	\mathcal{L}_{\text{MBM}} = \| \hat{\mathbf{F}}_t - \mathbf{F}^*_t \|_1.
\end{equation}
This loss is applied both during self-supervised pretraining and as an auxiliary term during downstream fine-tuning.

To further enforce structural consistency across patches and bins, MBM includes contrastive and consistency objectives. 
Let \(\mathbf{f}_{i,t}\) denote the feature of patch \(i\) in bin \(t\). 
A contrastive loss encourages features from the same spatial location in neighboring or augmented bins to be closer than features from different locations:
\begin{equation}
	\mathcal{L}_{\text{contrastive}} = - \sum_{i,t} \log \frac{\exp(\text{sim}(\mathbf{f}_{i,t}, \mathbf{f}_{i,t'}^+)/\tau)}{\sum_{j,t''} \exp(\text{sim}(\mathbf{f}_{i,t}, \mathbf{f}_{j,t''})/\tau)},
\end{equation}
where \(\mathbf{f}_{i,t'}^+\) is a positive feature from the same spatial location in a neighboring or augmented bin, \(\text{sim}(\cdot,\cdot)\) denotes a similarity metric (e.g., cosine similarity), and \(\tau\) is a temperature.

A consistency loss ensures stability of the predictions under augmentations:
\begin{equation}
	\mathcal{L}_{\text{consistency}} = \sum_{i,t} \| \hat{\mathbf{f}}_{i,t} - \hat{\mathbf{f}}_{i,t}^{\text{aug}} \|_2^2,
\end{equation}
where \(\hat{\mathbf{f}}_{i,t}^{\text{aug}}\) is the predicted feature for an augmented input.

\subsection{Training Loss}

We adopt a two-stage training strategy to leverage both self-supervised temporal feature learning and task-specific adaptation. 

In the first stage, MBM pretraining, the temporal encoder learns inter-bin dynamics from large amounts of unlabeled data, enhancing the generalization ability of the learned temporal features. 
The overall pretraining objective combines the reconstruction, contrastive, and consistency losses:
\begin{equation}
	\mathcal{L}_{\text{MBM-total}}^{\text{pretrain}} = \mathcal{L}_{\text{MBM}} + \lambda_1 \mathcal{L}_{\text{contrastive}} + \lambda_2 \mathcal{L}_{\text{consistency}},
\end{equation}
where \(\lambda_1\) and \(\lambda_2\) balance the contributions of the auxiliary losses.

In the second stage, joint fine-tuning, MBM is retained as an auxiliary self-supervised term while the entire network, including the DUSt3R backbone, is trained on 3D reconstruction. 

The joint training objective is
\begin{equation}
	\mathcal{L}_{\text{joint}} = L_{\text{regr}} + L_{\text{conf}} + \alpha L_{\text{MBM}} + \lambda_1 L_{\text{contrastive}} + \lambda_2 L_{\text{consistency}},
\end{equation}
where \(L_{\text{regr}}\) and \(L_{\text{conf}}\) denote the 3D regression and confidence-aware losses from the DUSt3R backbone. The remaining terms are auxiliary objectives: \(L_{\text{MBM}}\) is the masked bin modeling loss for self-supervised temporal feature learning, \(L_{\text{contrastive}}\) enforces cross-view feature alignment, and \(L_{\text{consistency}}\) encourages temporal coherence. Hyperparameters \(\alpha, \lambda_1, \lambda_2\) balance the contributions of the auxiliary objectives.

This two-stage strategy allows the network to first build robust temporal representations from unlabeled data and then adapt them effectively to task-specific objectives while maintaining structural and temporal coherence.

\subsection{Downstream Applications}
The output of \methodname is a globally aligned 3D pointmap, which can be readily used for downstream tasks such as multi-frame 3D reconstruction, depth estimation, and camera pose estimation. For these applications, we adopt the same post-processing and global alignment procedures as in DUSt3R, leveraging the rich 3D content and pixel-to-3D correspondences provided by the predicted pointmaps. Importantly, all predictions of depth, pose, and point coordinates correspond to the center voxel bin of the input event voxel grid, ensuring temporal consistency across frames. This allows \methodname to support standard downstream tasks without additional modifications.

\section{Experiments}
\label{sec:exps}

To the best of our knowledge, \methodname is the first event-only, feed-forward framework capable of performing globally consistent 3D reconstruction, making direct comparison with prior work not straightforward.
Therefore, we conduct a comprehensive evaluation of our approach across multiple 3D perception tasks, including depth estimation, camera pose estimation, and 3D reconstruction. 
For each task, we compare with representative state-of-the-art baselines when possible, or provide qualitative analyses in cases where no direct counterpart exists.

\subsection{Experiment Setup}
\paragraph{Implementation Details.} 
We use ViT-Base models for the temporal encoder as well as the spatial encoder and decoder. The spatial encoder and decoder are initialized with pretrained weights from DUSt3R.
We fix the event chunk length to 50 ms and divide it into T temporal bins. While larger T provides finer temporal resolution, it increases memory and computational costs. After balancing performance and efficiency, we select T=5.
To efficiently capture inter-bin dependencies, the temporal encoder operates on non-overlapping spatial patches of size $8 \times 8$. Each patch is embedded with a convolutional layer into a $32$-dimensional feature space. Temporal attention is performed over $2$ layers with $4$ attention heads and an MLP ratio of $4.0$, while no stochastic depth is applied. 

To enhance temporal feature learning, MBM randomly masks one or more temporal bins and trains a lightweight reconstruction decoder to predict the missing features from the visible bins. The MBM loss is defined as an L1 reconstruction loss, augmented with a contrastive loss over spatial patches and a consistency loss under input augmentations. The contrastive loss uses a temperature $\tau=0.1$, and the auxiliary loss weights are set to $\lambda_1 = 1.0$ and $\lambda_2 = 1.0$. During joint fine-tuning, MBM is retained as an auxiliary loss with weight $\alpha$ to reinforce modeling of temporal dependencies while optimizing the primary DUSt3R reconstruction loss.

We adopt a two-stage training strategy. In the first stage, MBM pretraining is performed for $300$ epochs using the Adam optimizer with a learning rate of $1\times10^{-4}$, weight decay of $0.05$, and batch size of $4$ on $8$ NVIDIA RTX A100 GPUs. In the second stage, the entire network is jointly fine-tuned on event–depth–pose paired datasets for $300$ epochs using the Adam optimizer with an initial learning rate of $5\times10^{-5}$ and cosine decay scheduling. Gradient clipping with a maximum norm of $1.0$ is applied for stability. The full training process takes approximately $7$ days on $8$ GPUs.

\paragraph{Datasets.} 
To train and evaluate the proposed method, we adopt a two-stage training and evaluation strategy that leverages both synthetic and real-world event data. 

In the first stage, the network is pretrained using the proposed Masked Bin Modeling (MBM) strategy on a diverse collection of synthetic and unlabeled real-world event datasets. Specifically, we leverage large-scale dataset, i.e.,  MatrixCity\cite{li2023matrixcity} and TUM-VIE\cite{klenk2021tum}. Since MatrixCity does not contain event streams, we simulate them using the ESIM simulator\cite{Gehrig_2020_CVPR} based on the available RGB sequences. In contrast, TUM-VIE provides real-world event data, although without aligned depth supervision. Together, these datasets cover both synthetic and real-world environments, as well as indoor and outdoor scenarios, enabling the model to learn generalizable spatio-temporal representations across diverse event domains before being fine-tuned on downstream tasks.

In the second stage, the network undergoes joint training on the TartanAir and MVSEC\cite{zhu2018multivehicle} datasets. Similar to MatrixCity, event streams for TartanAir are simulated from the available RGB sequences using the ESIM simulator, while retaining the provided ground-truth depth and camera poses under controlled conditions.
In contrast, MVSEC offers real-world event data recordings captured by a DAVIS346 camera mounted on a car and a hexacopter in outdoor and indoor environments, respectively. This setup enables a comprehensive evaluation of the proposed approach in terms of both generalization capability and real-world applicability.

\subsection{Monocular Depth Estimation}
We evaluate monocular depth estimation on the TartanAirEvent and MVSEC datasets using the same standard metrics as DUSt3R and CUT3R\cite{wang2025continuous}, namely Absolute Relative Error (AbsRel~$\downarrow$) and the threshold accuracy $\delta < 1.25$~($\uparrow$). All evaluations are performed with the official CUT3R code to ensure consistent and comparable results.

Quantitative results in Table~\ref{tab:QuanDepth} demonstrate that Event3R consistently outperforms prior event-based depth estimation methods \cite{wang2024evggs,hidalgo2020learning,Hitzges25neurips,liu2022event}. Our approach achieves substantially lower AbsRel and markedly higher fine-grained accuracy, indicating stronger geometric reconstruction and more reliable depth predictions from purely event-driven cues. These improvements stem from the combination of our globally aligned pointmap representation and the Temporal Encoder, which together enable more stable and coherent spatio-temporal aggregation than existing baselines.

Qualitative comparisons in Figure~\ref{fig:QualDepth} further reinforce these findings. Event3R produces sharper object boundaries, cleaner planar regions, and more globally consistent scene structures, while competing approaches often exhibit depth discontinuities, local artifacts, or scale inconsistencies. Overall, the results highlight the robustness and generalization ability of Event3R across diverse environments, validating the effectiveness of our event-only 3D reconstruction framework.

\input{sec/fig_QualDepth}

\input{sec/table_QuanDepth}

\subsection{Camera Pose Estimation}

\input{sec/table_PoseEstimate}

We evaluate camera pose estimation on the TartanAir benchmark using Absolute Trajectory Error (ATE) computed with the Evo evaluation tool. Quantitative results are reported in Table~\ref{tab:PoseEstimate}. Event3R is compared against IncEventGS~\cite{huang2025inceventgs}, DEVO~\cite{DEVO}, and a two-stage e2vid~\cite{e2vid}+COLMAP~\cite{colmap} pipeline for event-based pose estimation.
As shown in Table~\ref{tab:PoseEstimate}, Event3R consistently achieves the lowest ATE across all tested sequences, demonstrating more accurate and stable trajectory estimation. These gains highlight the effectiveness of our globally aligned pointmap representation and Temporal Encoder in extracting reliable motion cues from asynchronous event streams.

\subsection{3D Reconstruction}

\begin{table}
    \centering
    \setlength{\tabcolsep}{5pt}
    \caption{\textbf{Quantitative Results for 3D Reconstruction.}}
    \vspace{0.0em}
    \resizebox{0.48\textwidth}{!}{%
    \begin{tabular}{@{}cc@{\hspace{3pt}}c@{\hspace{3pt}}cc@{\hspace{3pt}}c@{\hspace{3pt}}c@{}}
        \specialrule{0.12em}{1pt}{1pt}
        Method & \multicolumn{3}{c}{MVSEC} & \multicolumn{3}{c}{TartanAir} \\
        \cmidrule(lr){2-4} \cmidrule(lr){5-7}
        & Acc $\downarrow$ & Comp$ \downarrow$ & NC $\uparrow$ 
        & Acc $\downarrow$ & Comp$ \downarrow$ & NC $\uparrow$ \\
        \specialrule{0.05em}{1pt}{1pt}
        Ours & \textbf{0.4915} & \textbf{0.6132} & \textbf{0.7569} 
             & \textbf{0.1097} & \textbf{0.1382} & \textbf{0.8886}\\
        DepthAnyEvent & 0.5342 & 0.7539 & 0.6195 
                      & 0.5110 & 0.5764  & 0.6576 \\
        E2Depth & 0.7987 & 0.6532 & 0.6332 
                & 0.6891 & 0.4952 & 0.4950 \\
        IncEventGS & 0.6879 & 0.6231 & 0.5332 
                & 0.3350 & 0.3609 & 0.6236 \\
        \specialrule{0.12em}{1pt}{1pt}
    \end{tabular}%
    }
    \label{tab:pt3d_recon}
    \vspace{-2.5em}
\end{table}

After training, we perform inference on several two-view pairs as well as multi-view sequences ($N>2$) from different datasets, and globally align the estimated poses to evaluate the resulting 3D reconstructions. We present qualitative results for both the two-view and multi-view (e.g., five-view) settings. Due to space limitations, qualitative results are provided in the supplementary video. 
 We also present quantitative results on 3D reconstruction in \tabnref{tab:pt3d_recon}. Since Event3R is a new feed-forward event-based 3D reconstruction method, there is no direct baseline for comparison. To further evaluate our approach, we compare it with IncEventGS and a self-implemented two-stage pipeline (DepthAnything followed by ground-truth pose projection). The results further demonstrate the advantages of our method.

\subsection{Ablation Study}

We evaluate the impact of the Temporal Encoder and Masked Bin Modeling (MBM)—including its self-supervised pretraining and auxiliary contrastive and consistency losses—on depth and pose estimation. Removing the Temporal Encoder (i.e., feeding the voxelized events directly into the Spatial Encoder without temporal aggregation) leads to a substantial degradation in both depth (AbsRel: 0.0514 $\rightarrow$ 0.1187) and pose (ATE: 0.1524 $\rightarrow$ 0.4381 cm), confirming the necessity of patch-level temporal aggregation for capturing fine-grained motion cues.

MBM pretraining shows particularly strong impact: disabling pretraining increases depth and pose errors significantly (AbsRel: 0.0514 $\rightarrow$ 0.0825, ATE: 0.1524 $\rightarrow$ 0.3095 cm), while ablating its contrastive or consistency objectives further degrades performance. These results highlight that MBM not only enhances module's temporal feature learning during supervised training but also provides a robust initialization, greatly reducing the reliance on labeled data.

Overall, TE and MBM—including pretraining and auxiliary objectives—jointly enable robust, temporally coherent spatio-temporal representations, resulting in more accurate event-based depth and pose estimation.

\begin{table}[!t]
	\centering
	\small
	\caption{\textbf{Ablation study of Event3R components.} Both TE and MBM—especially self-supervised pretraining—are crucial; disabling MBM pretraining or its contrastive/consistency objectives leads to performance drops.}
	\begin{tabular}{l|cc|c}
		\specialrule{0.12em}{1pt}{1pt}
		Model & AbsRel $\downarrow$ & $\delta<1.25$ $\uparrow$ & ATE $\downarrow$ \\
		\specialrule{0.05em}{1pt}{1pt}
		Full Event3R & \textbf{0.0514} & \textbf{0.970} & \textbf{0.1524} \\
		w/o TE & 0.1187 & 0.892 & 0.4381 \\
		w/o MBM pretrain & 0.0825 & 0.925 & 0.3095 \\
		MBM w/o contrastive & 0.0621 & 0.961 & 0.1792 \\
		MBM w/o consistency & 0.0598 & 0.965 & 0.1675 \\
		\specialrule{0.12em}{1pt}{1pt}
	\end{tabular}
	\label{tab:ablation}
    \vspace{-2.5em}
\end{table}

\section{Conclusion}
\label{sec:conclusion}
\vspace{-0.14cm}

We present \methodname, a new feed-forward, event-only framework for globally consistent 3D reconstruction from asynchronous event streams. By integrating a patch-level Temporal Encoder with a spatial backbone, our method captures rich spatio-temporal features encoding fine-grained motion and geometric cues. Masked Bin Modeling further strengthens temporal learning, enabling self-supervised pretraining, reducing dependence on labeled data, and enforcing inter-bin structural consistency. Extensive experiments on synthetic and real-world datasets show that Event3R produces sharper depth, more coherent 3D pointmaps, and more reliable camera trajectories, validating its robustness and generalization. Event3R enables efficient, scalable, and label-efficient 3D perception from event sensors, with broad applications in robotics, autonomous navigation, and real-time scene understanding.

\newlength{\bibitemsep}\setlength{\bibitemsep}{0.0\baselineskip}
\newlength{\bibparskip}\setlength{\bibparskip}{0pt}
\let\oldthebibliography\thebibliography
\renewcommand\thebibliography[1]{%
	\oldthebibliography{#1}%
	\setlength{\parskip}{\bibitemsep}%
	\setlength{\itemsep}{\bibparskip}%
}
\bibliography{IROS2026}

\input{sec/X_suppl}

\end{document}

%% file: preamble.tex
\usepackage{microtype}

\input{shortcuts}

\makeatletter
\DeclareRobustCommand\onedot{\futurelet\@let@token\@onedot}
\def\@onedot{\ifx\@let@token.\else.\null\fi\xspace}
 
\def\ie{\emph{i.e}\onedot}

\def\etal{\emph{et al}\onedot}
\makeatother

\newcommand{\methodname}{Event3R\xspace}

%% file: shortcuts.tex
\newcommand{\tabnref}[1]{Table~\ref{#1}}

\makeatletter
\DeclareRobustCommand\onedot{\futurelet\@let@token\@onedot}
\def\@onedot{\ifx\@let@token.\else.\null\fi\xspace}
 
\def\ie{i.e\onedot}

\def\etal{et~al\onedot}

\makeatother

\newcommand{\PAR}[1]{\vspace{0.1cm}\noindent{\bf #1} }


%% file: sec/fig_QualDepth.tex
\begin{figure*}
    \centering
    \setlength{\tabcolsep}{1pt}
	\begin{tabular}{p{8.2pt}cccc}
		& {Ground Truth} & {Event3R} & {EvGGS} & {E2Depth}  \\
		\rotatebox[origin=c]{90}{\small {Synthetic}} & 
		\parbox[c]{0.20\linewidth}{\vspace{1pt}\centering\includegraphics[width=\linewidth]{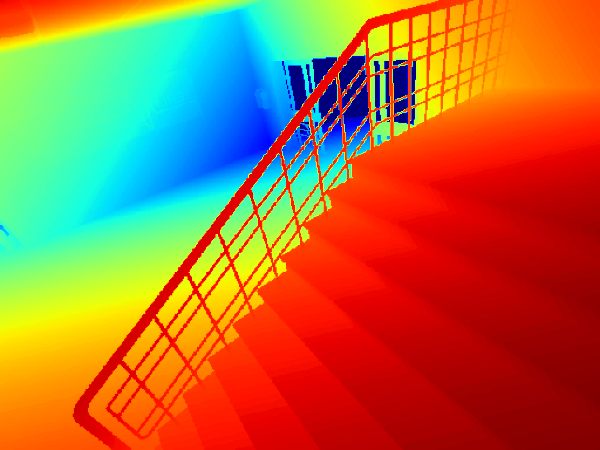}} &
		\parbox[c]{0.20\linewidth}{\vspace{1pt}\centering\includegraphics[width=\linewidth]{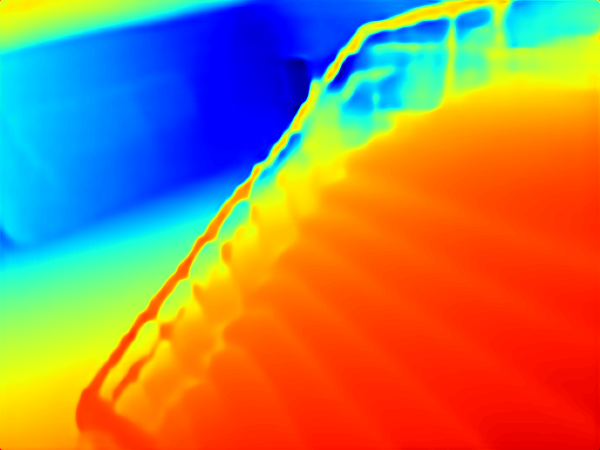}} &
		\parbox[c]{0.20\linewidth}{\vspace{1pt}\centering\includegraphics[width=\linewidth]{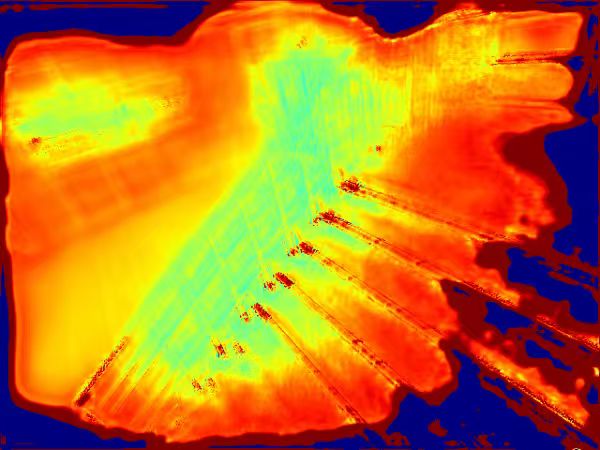}}&
		\parbox[c]{0.20\linewidth}{\vspace{1pt}\centering\includegraphics[width=\linewidth]{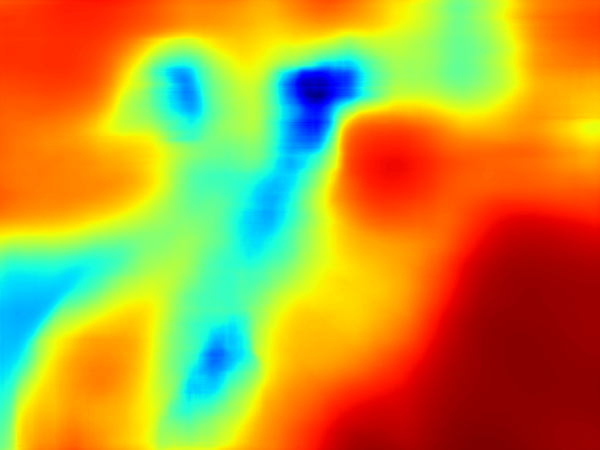}}
        \\
		
        \rotatebox[origin=c]{90}{\small {Synthetic}} & 
		\parbox[c]{0.20\linewidth}{\vspace{3pt}\centering\includegraphics[width=\linewidth]{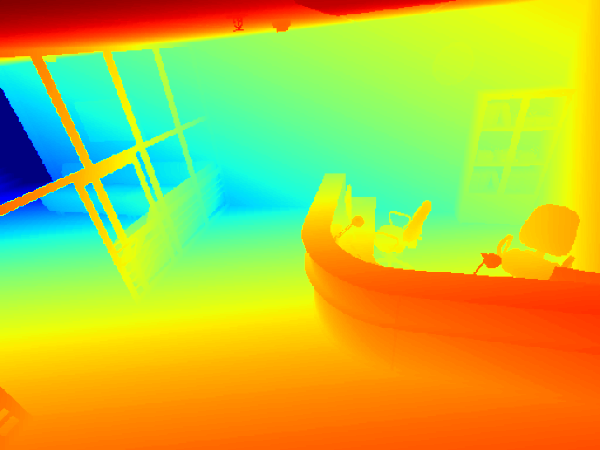}} &
		\parbox[c]{0.20\linewidth}{\vspace{3pt}\centering\includegraphics[width=\linewidth]{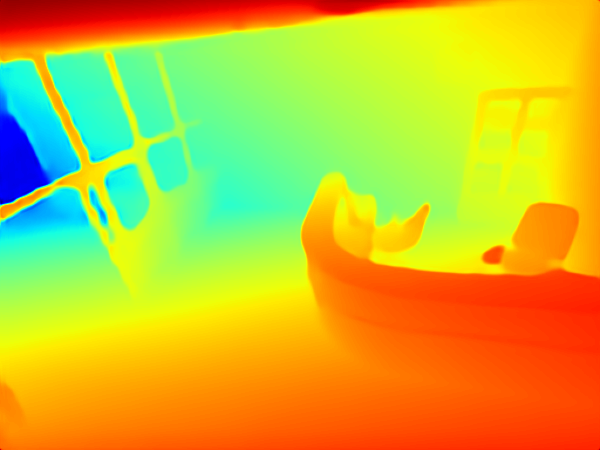}} &
		\parbox[c]{0.20\linewidth}{\vspace{3pt}\centering\includegraphics[width=\linewidth]{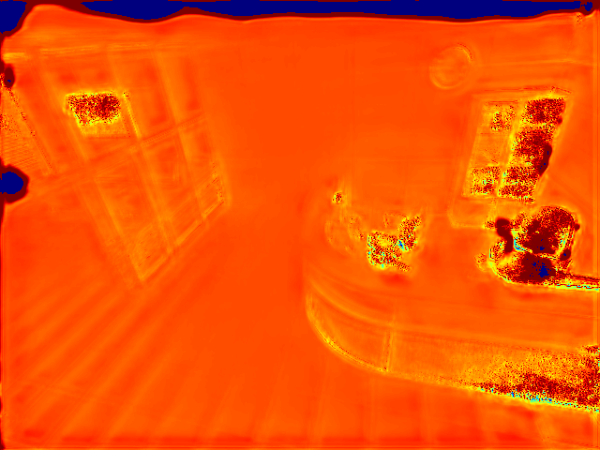}} &
		\parbox[c]{0.20\linewidth}{\vspace{3pt}\centering\includegraphics[width=\linewidth]{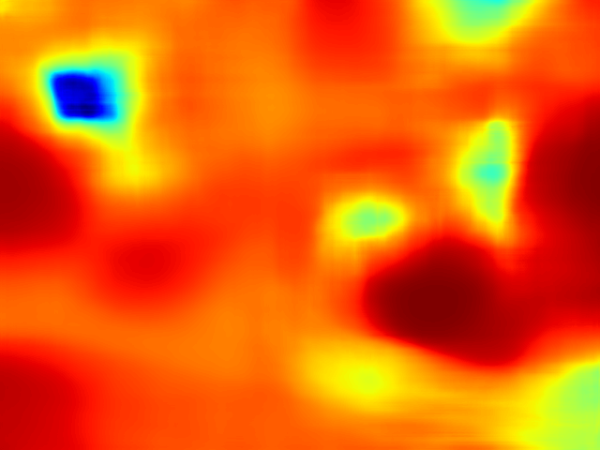}}\\
		
        \rotatebox[origin=c]{90}{\small {Synthetic}} & 
		\parbox[c]{0.20\linewidth}{\vspace{3pt}\centering\includegraphics[width=\linewidth]{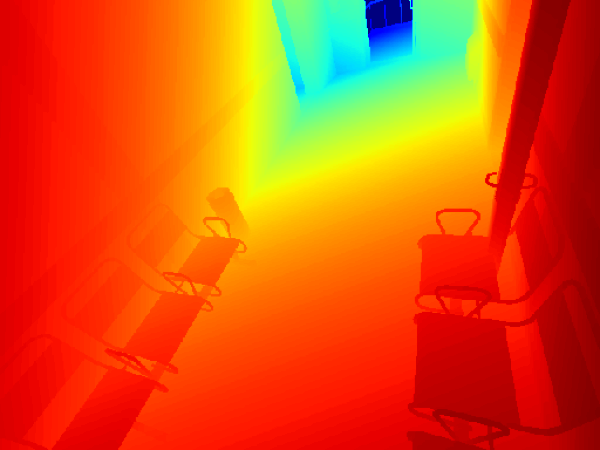}} &
		\parbox[c]{0.20\linewidth}{\vspace{3pt}\centering\includegraphics[width=\linewidth]{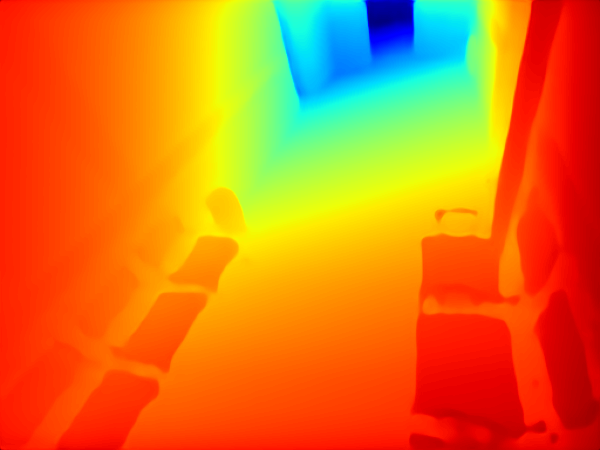}} &
		\parbox[c]{0.20\linewidth}{\vspace{3pt}\centering\includegraphics[width=\linewidth]{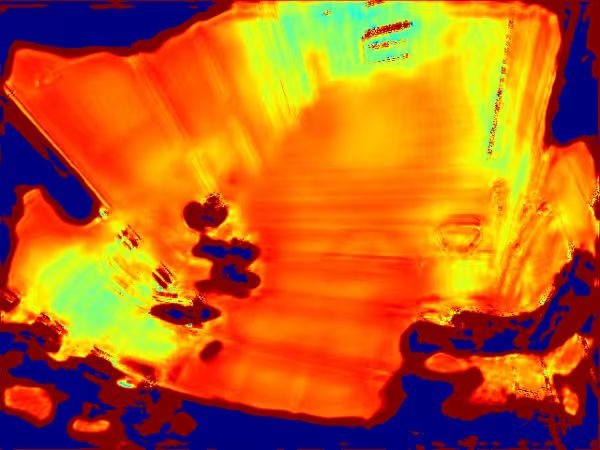}}&
		\parbox[c]{0.20\linewidth}{\vspace{3pt}\centering\includegraphics[width=\linewidth]{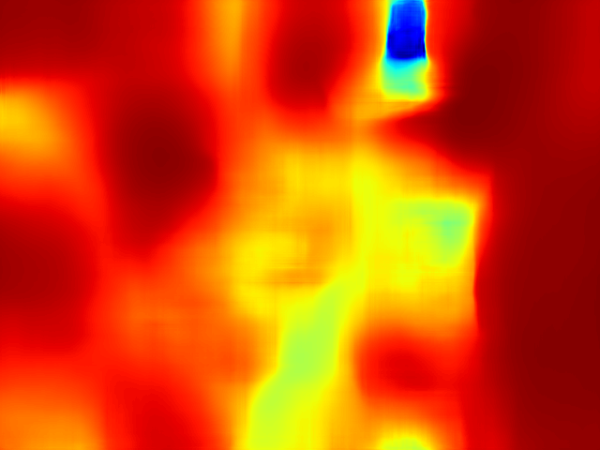}} \\
		
        \rotatebox[origin=c]{90}{\small {Synthetic}} & 
		\parbox[c]{0.20\linewidth}{\vspace{3pt}\centering\includegraphics[width=\linewidth]{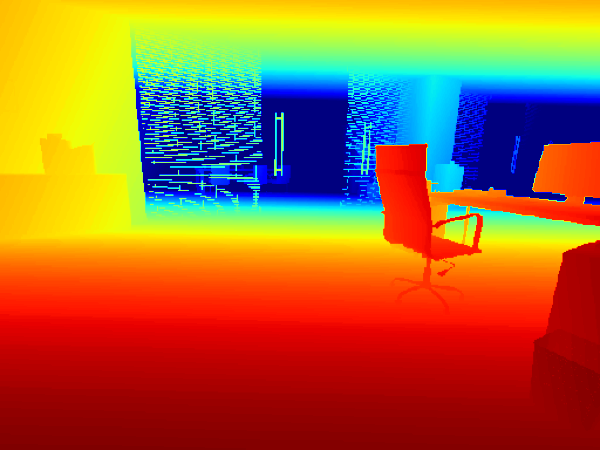}} &
		\parbox[c]{0.20\linewidth}{\vspace{3pt}\centering\includegraphics[width=\linewidth]{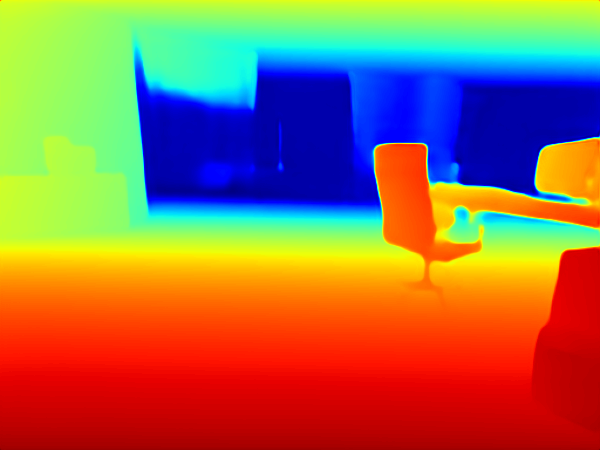}} &
		\parbox[c]{0.20\linewidth}{\vspace{3pt}\centering\includegraphics[width=\linewidth]{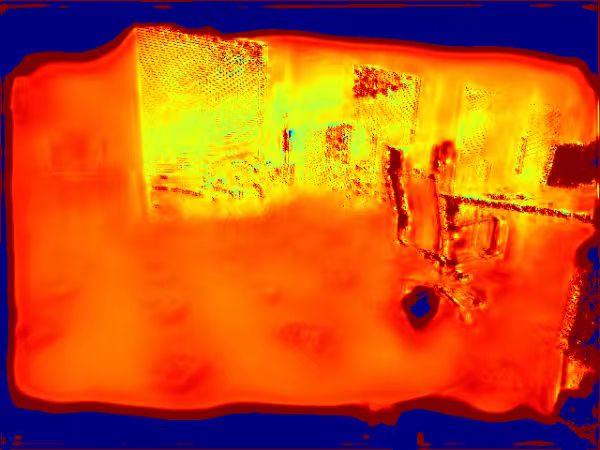}}&
		\parbox[c]{0.20\linewidth}{\vspace{3pt}\centering\includegraphics[width=\linewidth]{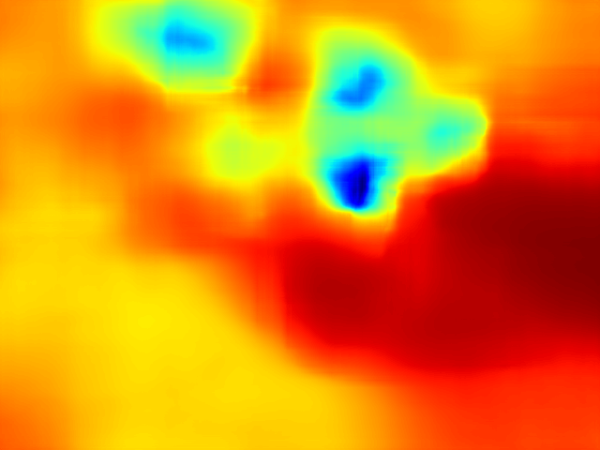}} \\
		
        \rotatebox[origin=c]{90}{\small {Real-world}} & 
		\parbox[c]{0.20\linewidth}{\vspace{3pt}\centering\includegraphics[width=\linewidth]{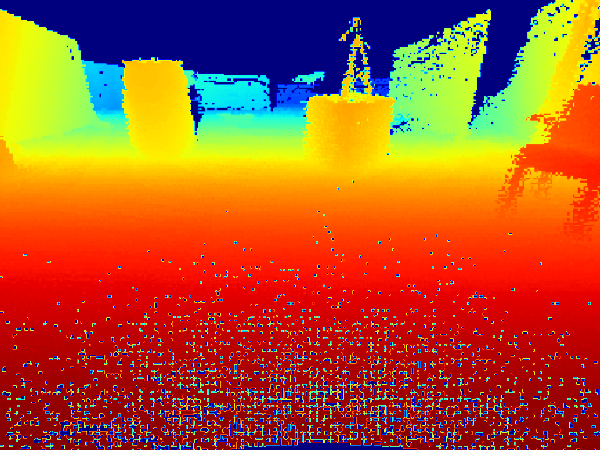}} &
		\parbox[c]{0.20\linewidth}{\vspace{3pt}\centering\includegraphics[width=\linewidth]{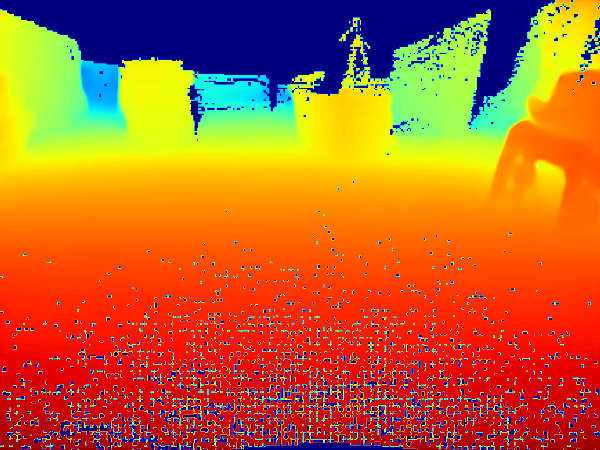}} &
		\parbox[c]{0.20\linewidth}{\vspace{3pt}\centering\includegraphics[width=\linewidth]{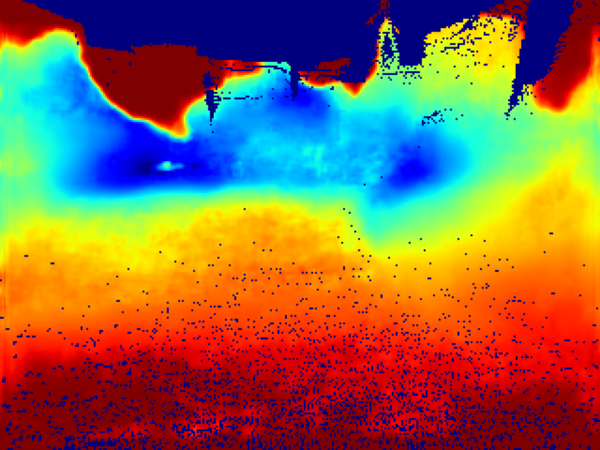}}&
		\parbox[c]{0.20\linewidth}{\vspace{3pt}\centering\includegraphics[width=\linewidth]{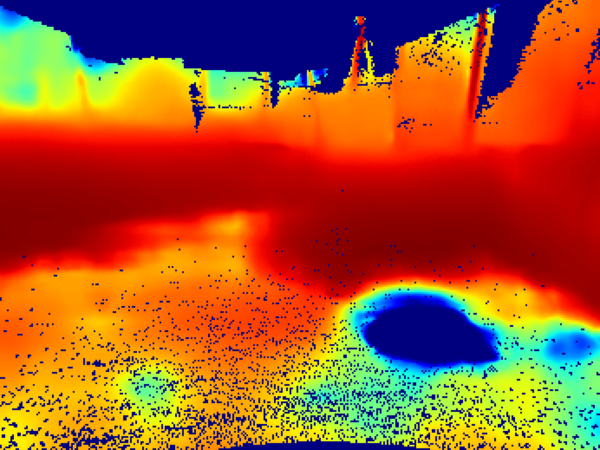}} \\

        \rotatebox[origin=c]{90}{\small {Real-world}} & 
		\parbox[c]{0.20\linewidth}{\vspace{3pt}\centering\includegraphics[width=\linewidth]{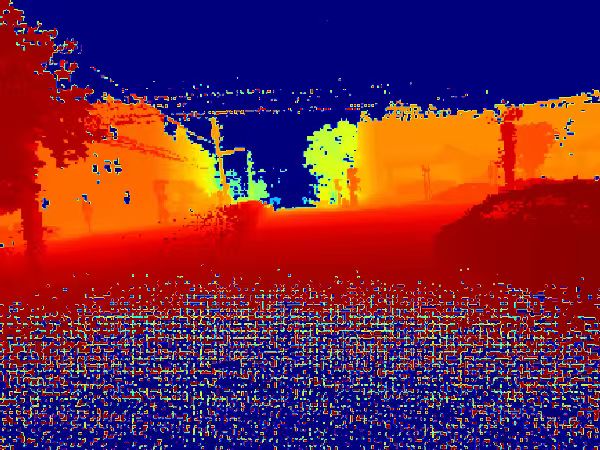}} &
		\parbox[c]{0.20\linewidth}{\vspace{3pt}\centering\includegraphics[width=\linewidth]{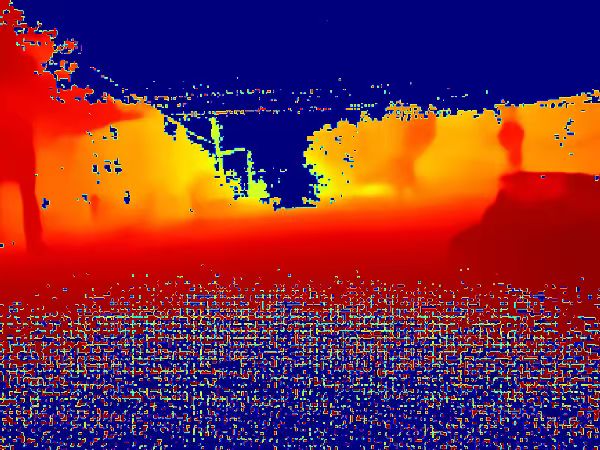}} &
		\parbox[c]{0.20\linewidth}{\vspace{3pt}\centering\includegraphics[width=\linewidth]{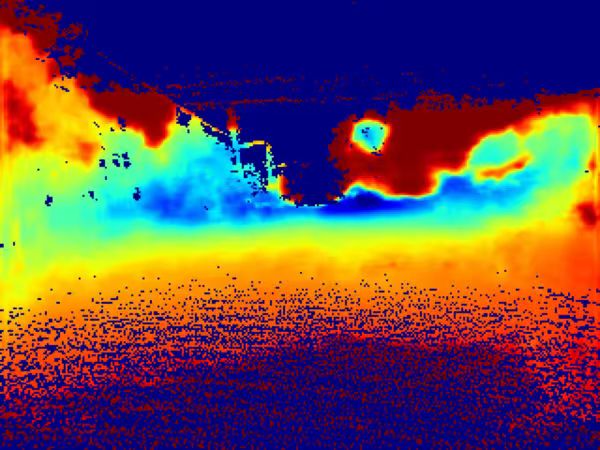}}&
		\parbox[c]{0.20\linewidth}{\vspace{3pt}\centering\includegraphics[width=\linewidth]{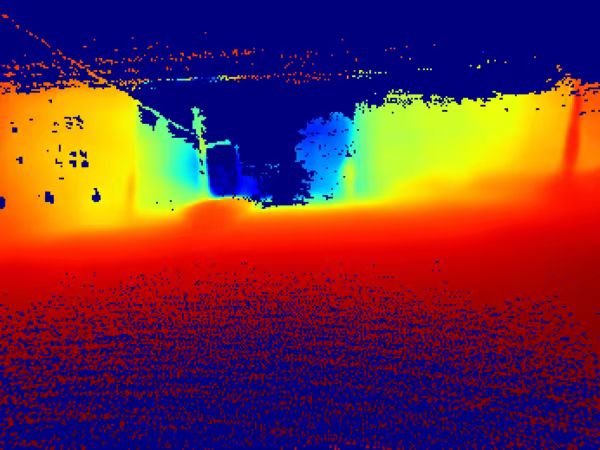}} \\
		
	\end{tabular}
    \vspace{-5pt}
    \caption{\textbf{Qualitative comparison of depth estimation results on both synthetic and real-world datasets}. For better visualization, the grayscale depth maps are colorized using the 'jet' colormap.}
    \label{fig:QualDepth}
    \vspace{-1.2em}
\end{figure*}

%% file: sec/table_QuanDepth.tex
\begin{table} 
    \centering
    \small
    \setlength{\tabcolsep}{3pt}
    \caption{\textbf{Quantitative depth estimation results on event-based datasets.} Comparison of our method against state-of-the-art event-based depth estimation approaches on TartanAir and MVSEC outdoor datasets.}
    \begin{tabular}{ccccc}
        \specialrule{0.12em}{1pt}{1pt}
        Method & \multicolumn{2}{c}{TartanAir} & \multicolumn{2}{c}{MVSEC} \\
        \cmidrule(lr){2-3} \cmidrule(lr){4-5}
        & Abs Rel $\downarrow$ & $\delta < 1.25 \uparrow$ & Abs Rel $\downarrow$ & $\delta < 1.25 \uparrow$ \\
        \specialrule{0.05em}{1pt}{1pt}
        Ours & \textbf{0.0514} & \textbf{0.9700} & \textbf{0.1805} & \textbf{0.7838} \\
        EvGGS & 0.4553 & 0.5420 & 0.1972 & 0.7138 \\
        E2Depth & 0.7923 & 0.2484 & 0.3763 & 0.5073 \\
        DepthAnyEvent & 0.2751&0.6093 & 0.4100& 0.3783 \\
        DERD-Net & 0.5651 & 0.2488 & 0.4367 & 0.2288 \\
        EReFormer & 0.2931 & 0.5323 & 0.4986 & 0.3799 \\
        \specialrule{0.12em}{1pt}{1pt}
    \end{tabular}
    \label{tab:QuanDepth}
    \vspace{-4.5em}
\end{table}

%% file: sec/table_PoseEstimate.tex
\begin{table} 
    \centering
    \small
    \caption{{\textbf{ Quantitative pose estimation results on event-based datasets(ATE, cm).}} Comparison of our method against state-of-the-art event-based pose estimation approaches on TartanAir dataset. 
    S1: Hospital\_Easy\_P004, S2: Hospital\_Easy\_P008, S3: Hospital\_Easy\_P011, S4: Office\_Hard\_P006, S5: Office\_Easy\_P000. “-” indicates that the experiment failed on this scene.}
    \begin{tabular}{c|ccccc}
        \specialrule{0.12em}{1pt}{1pt}
        Method & S1 & S2 & S3 & S4 & S5 \\
        \specialrule{0.05em}{1pt}{1pt}
        Ours & \textbf{0.402} & \textbf{0.152} & \textbf{0.116} & \textbf{0.545} & \textbf{0.293} \\
        DEVO & 1.25  & 0.457 & 0.891 & 2.02  & 0.861 \\
        E2Vid + Colmap & - & 0.550 & 0.828 & 1.87  & 1.18  \\
        IncEventGS & 1.16 & 0.307 & 0.261 & 1.56  & 1.03  \\
        \specialrule{0.12em}{1pt}{1pt}
    \end{tabular}
    \label{tab:PoseEstimate}
\end{table}

%% file: sec/X_suppl.tex
\clearpage
\setcounter{page}{1}

\begin{figure*}[!t]
    \centering
    \large
    \textbf{Event3R: Asynchronous-to-Global 3D Reconstruction from Event Camera via Spatial-Temporal Feature Aggregation}\\
    \vspace{0.3em}Supplementary Material \\
    \vspace{0.8em}
    \normalsize
    \setlength{\tabcolsep}{1pt}
    \begin{tabular}{cc}
        \small{View 1} & \small{View 2} \\
        \parbox[c]{0.45\linewidth}{\vspace{1pt}\centering\includegraphics[width=\linewidth]{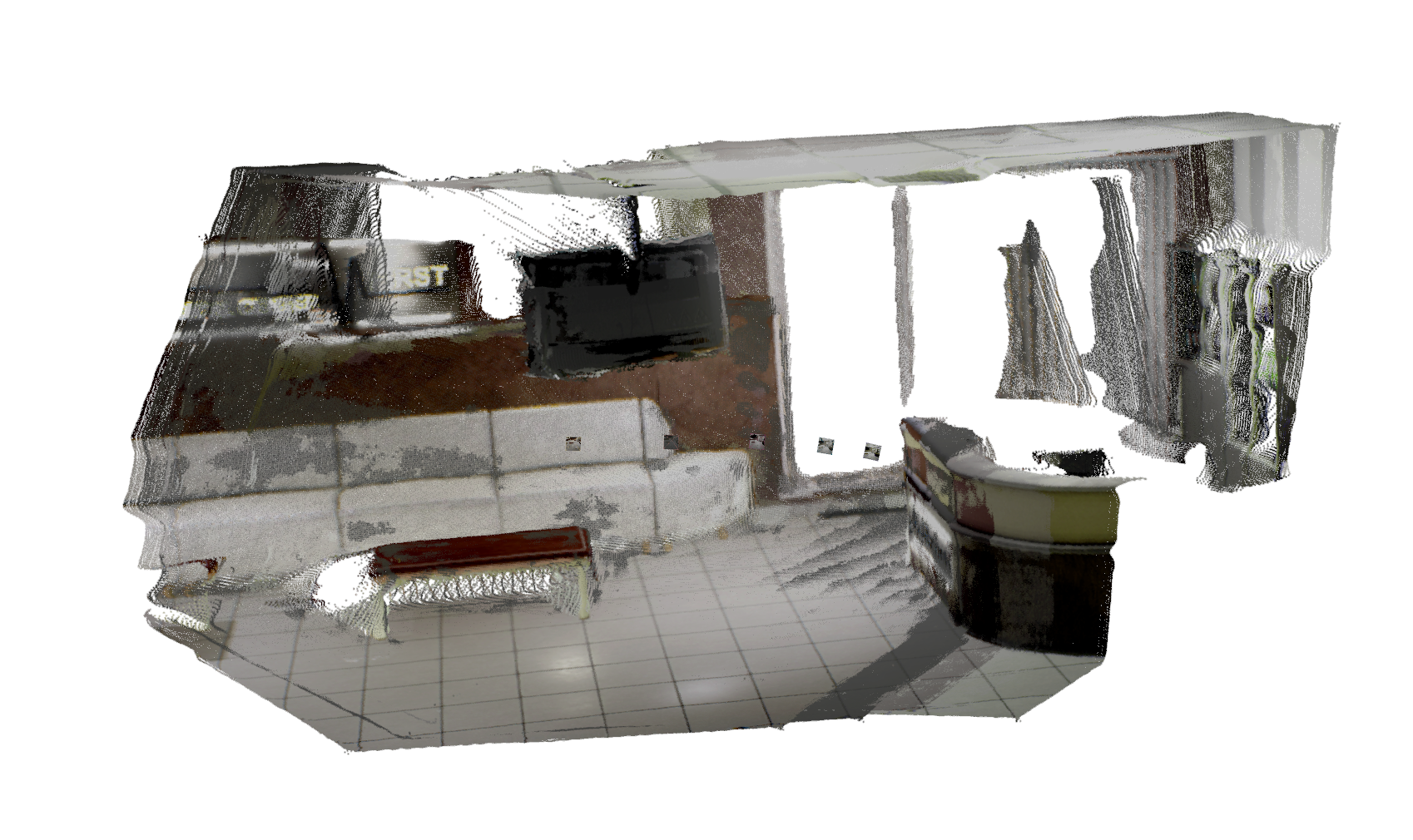}} &
        \parbox[c]{0.45\linewidth}{\vspace{3pt}\centering\includegraphics[width=\linewidth]{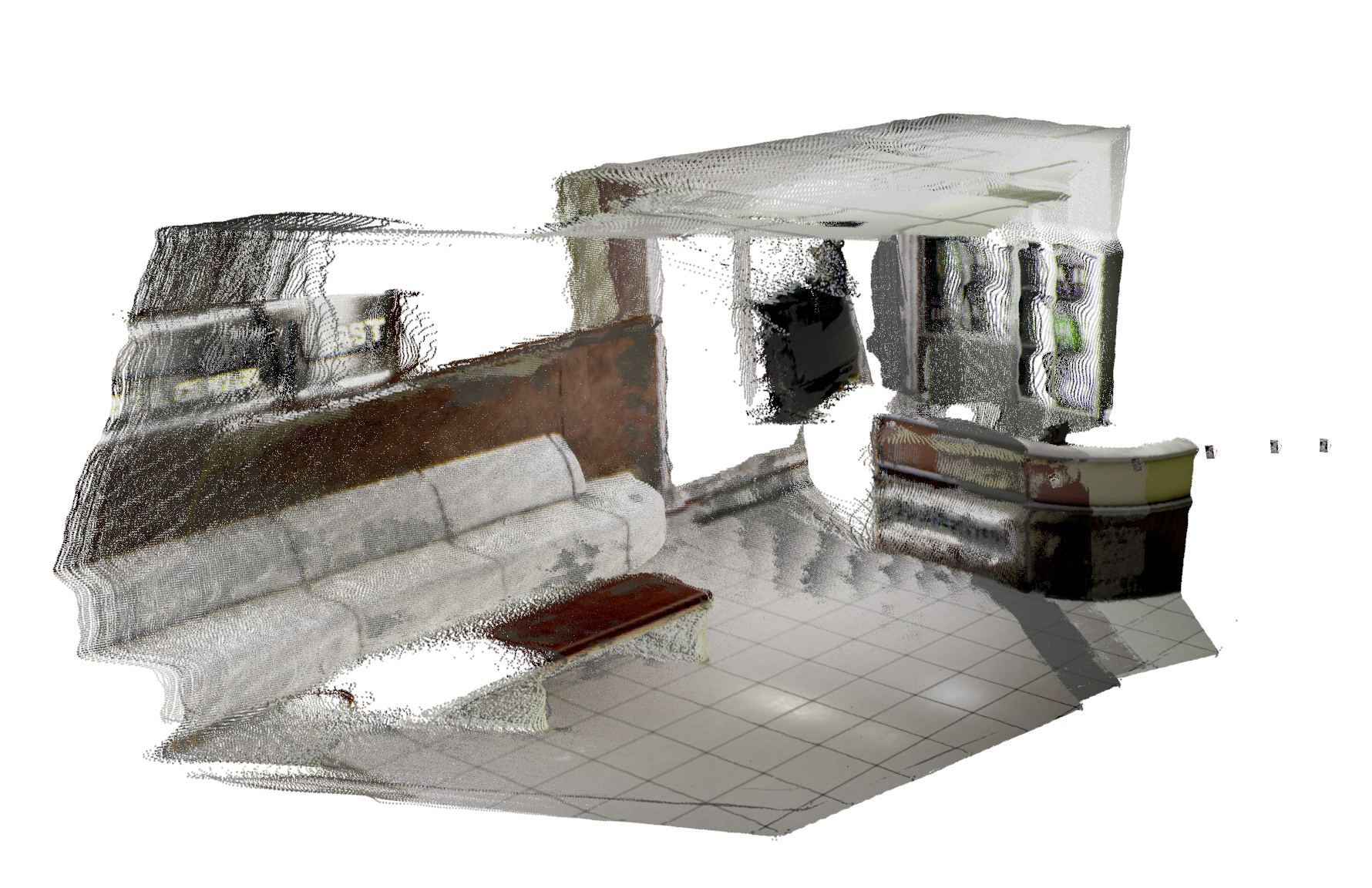}} \\

        \parbox[c]{0.45\linewidth}{\vspace{1pt}\centering\includegraphics[width=\linewidth]{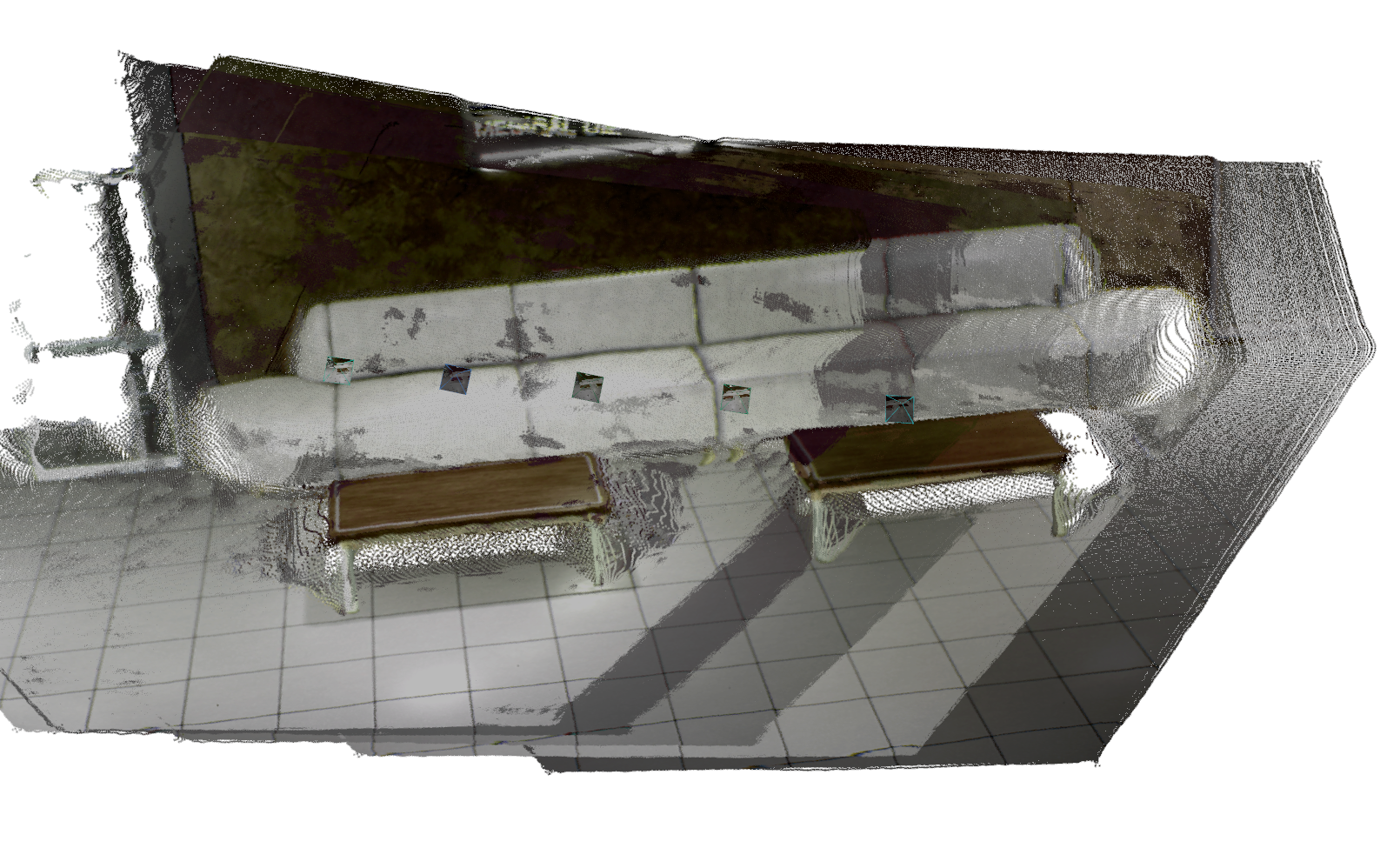}} &
        \parbox[c]{0.45\linewidth}{\vspace{3pt}\centering\includegraphics[width=\linewidth]{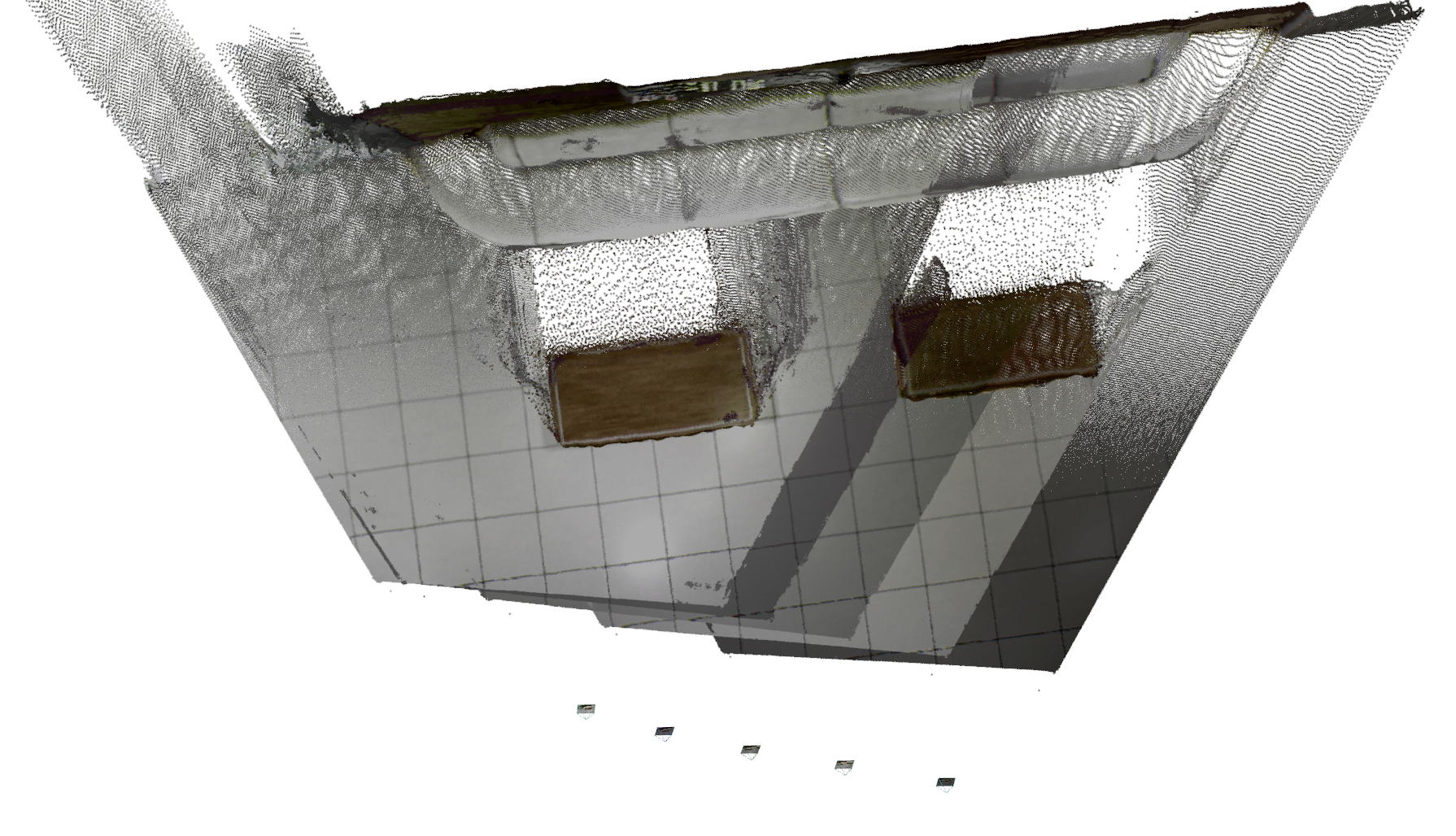}} \\

    \end{tabular}
    \vspace{-3pt}
    \caption{\textbf{Global multi-view 3D reconstruction results.} We demonstrate the capability of Event3R to reconstruct globally consistent 3D scenes from multiple views ($N=5$). The estimated camera trajectories are visualized as frustums alongside the point clouds, demonstrating accurate global alignment.}
    \label{fig:suppl_Multiview}
    \vspace{-1em}
\end{figure*}

\section{Additional Qualitative 3D Reconstructions}

\label{sec:additional_demo_3D}

In the supplementary material, we include an expanded collection of qualitative results to illustrate the 3D reconstruction performance of our method under different view configurations. We provide additional visualizations for both the two-view setting and multi-view sequences. 

For visualization, the reconstructed point clouds are colorized using the RGB image corresponding to the center voxel bin. No RGB data is involved during training or inference.

\paragraph{Two-view Reconstruction} In the two-view setting, the pointmaps predicted by our \methodname model are already expressed in a consistent global coordinate frame, so no additional post-processing or global alignment is required. Reconstruction results on synthetic datasets are shown in Figure~\ref{fig:suppl_Twoview_syn}.

\paragraph{Multi-view Processing}
To reconstruct a consistent 3D scene from $N$ views using only event voxel grids, we follow the global alignment framework of DUSt3R. We construct a connectivity graph where each node corresponds to a view and each edge denotes a pair with sufficient overlap. For every edge, the network predicts pairwise pointmaps with confidence maps, which are then fused through a global optimization that minimizes the weighted 3D reprojection error. This jointly recovers the absolute camera poses, intrinsic parameters, and depthmaps for the entire sequence. The reconstruction results, presented in Figure~\ref{fig:suppl_Multiview}, illustrate accurate global alignment and consistent geometry across all views.

\input{sec/fig_twoview_syn}

%% file: sec/fig_twoview_syn.tex
\begin{figure*}[!h]
    \centering
    \setlength{\tabcolsep}{1pt}
    \begin{tabular}{cc}
        \small{View 1} & \small{View 2} \\
        \parbox[c]{0.45\linewidth}{\vspace{1pt}\centering\includegraphics[width=\linewidth]{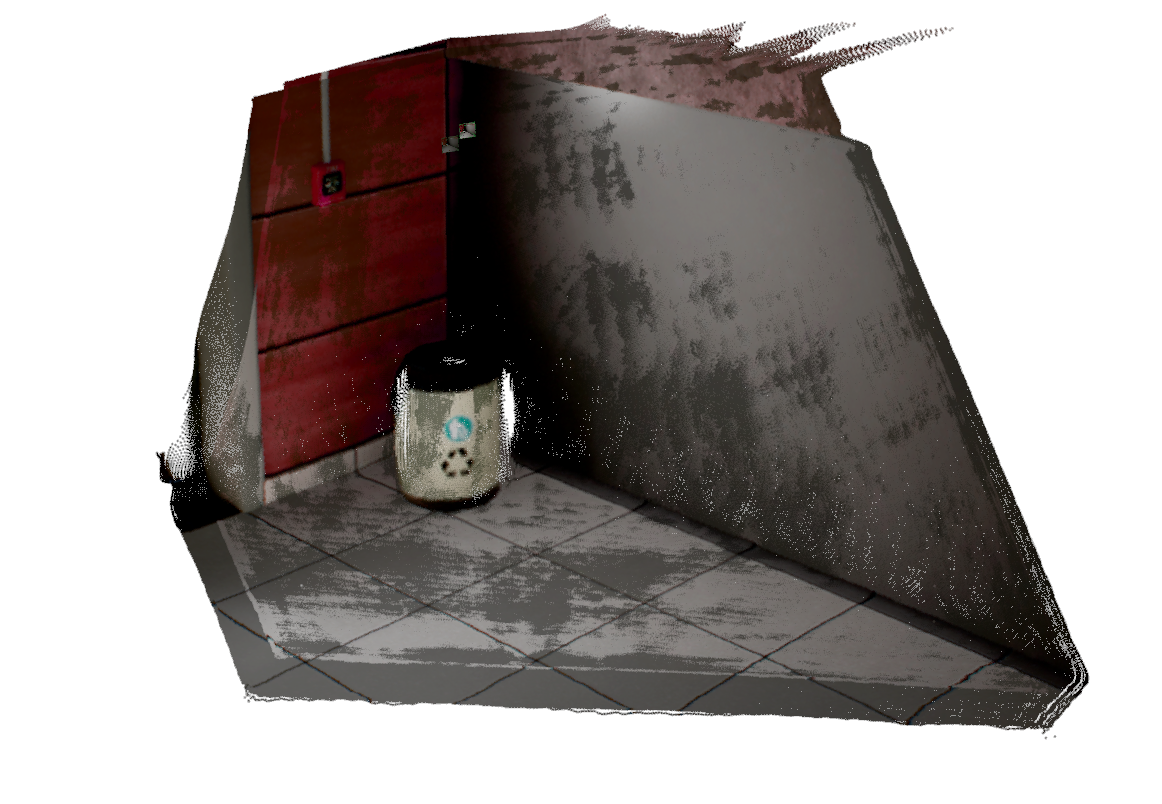}} &
        \parbox[c]{0.45\linewidth}{\vspace{3pt}\centering\includegraphics[width=\linewidth]{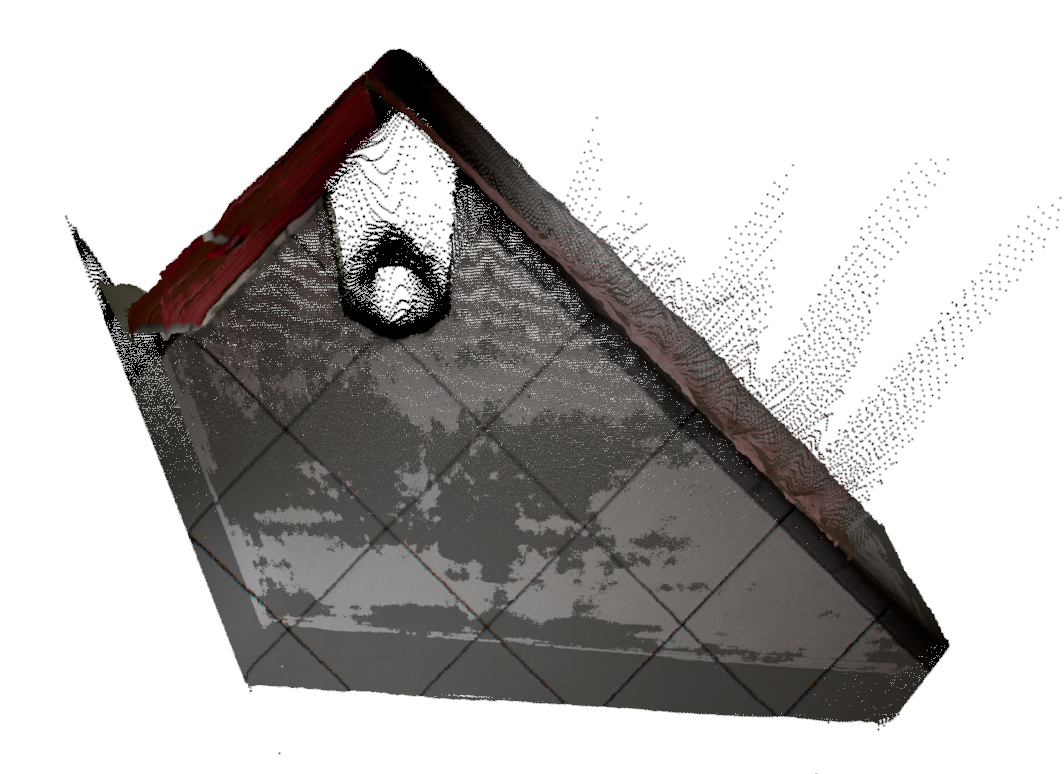}} \\

        \parbox[c]{0.45\linewidth}{\vspace{1pt}\centering\includegraphics[width=\linewidth]{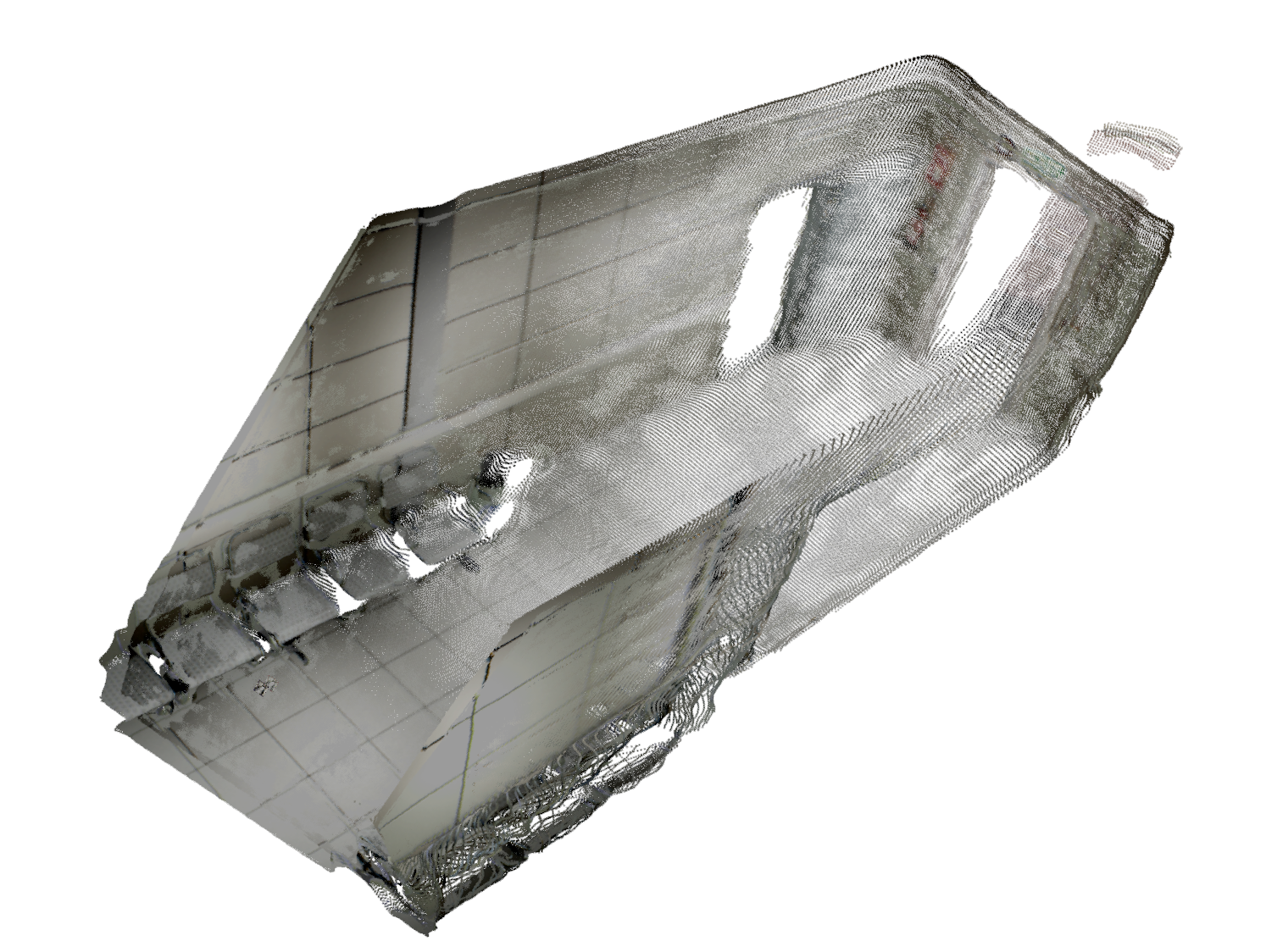}} &
        \parbox[c]{0.45\linewidth}{\vspace{3pt}\centering\includegraphics[width=\linewidth]{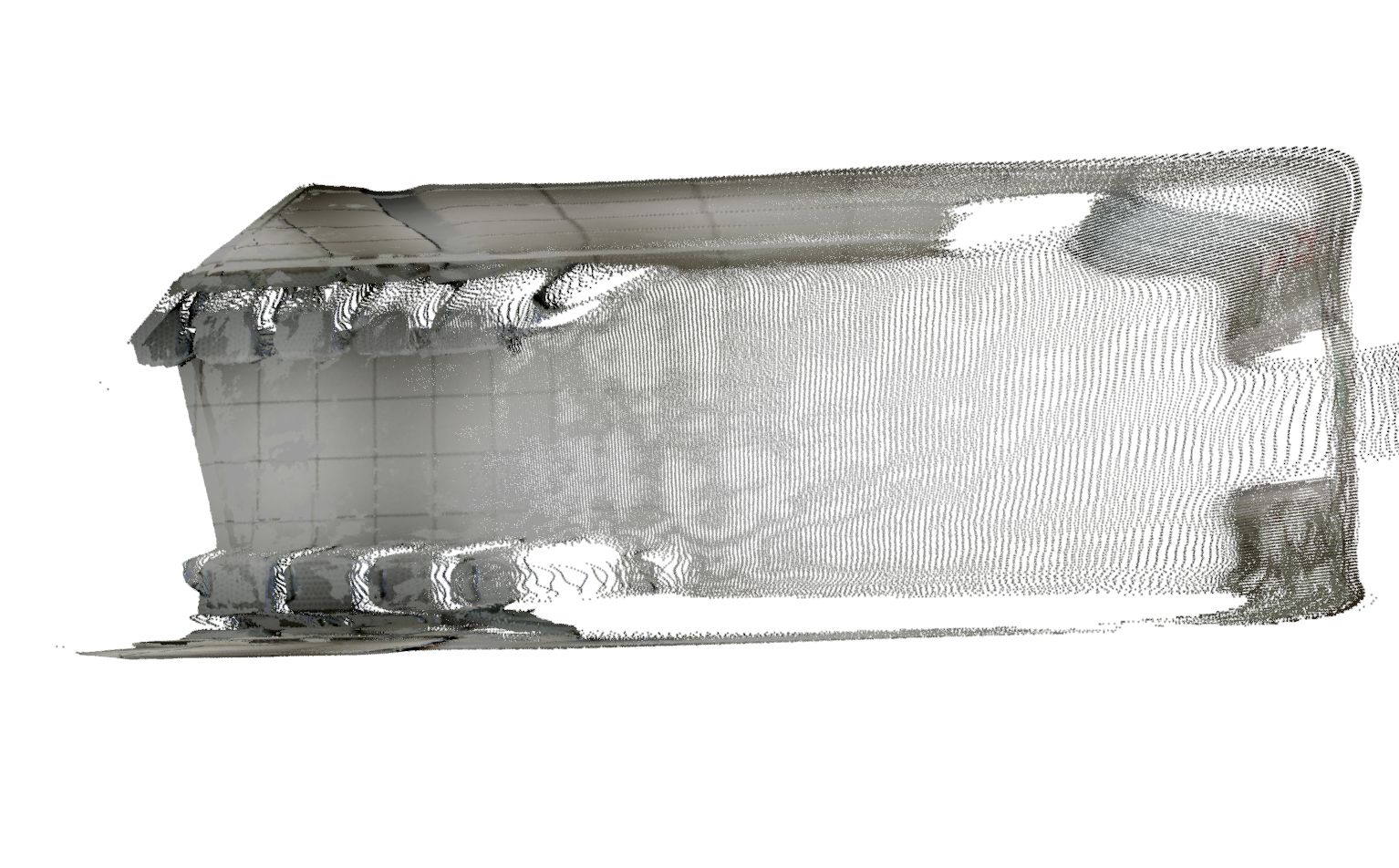}} \\

        \parbox[c]{0.45\linewidth}{\vspace{1pt}\centering\includegraphics[width=\linewidth]{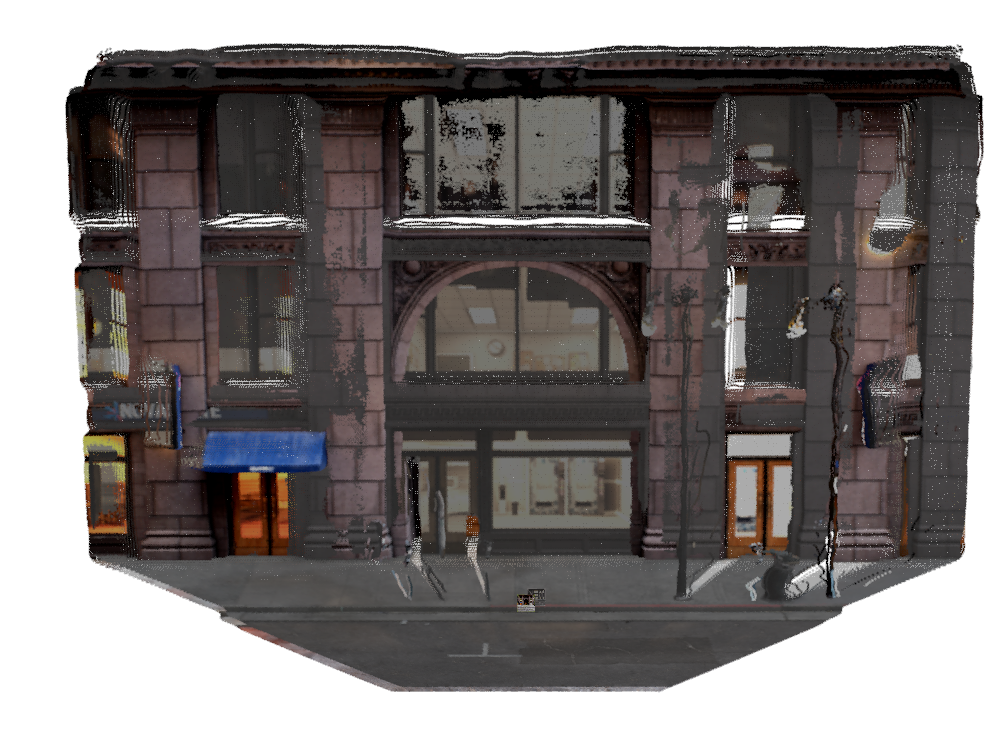}} &
        \parbox[c]{0.45\linewidth}{\vspace{3pt}\centering\includegraphics[width=\linewidth]{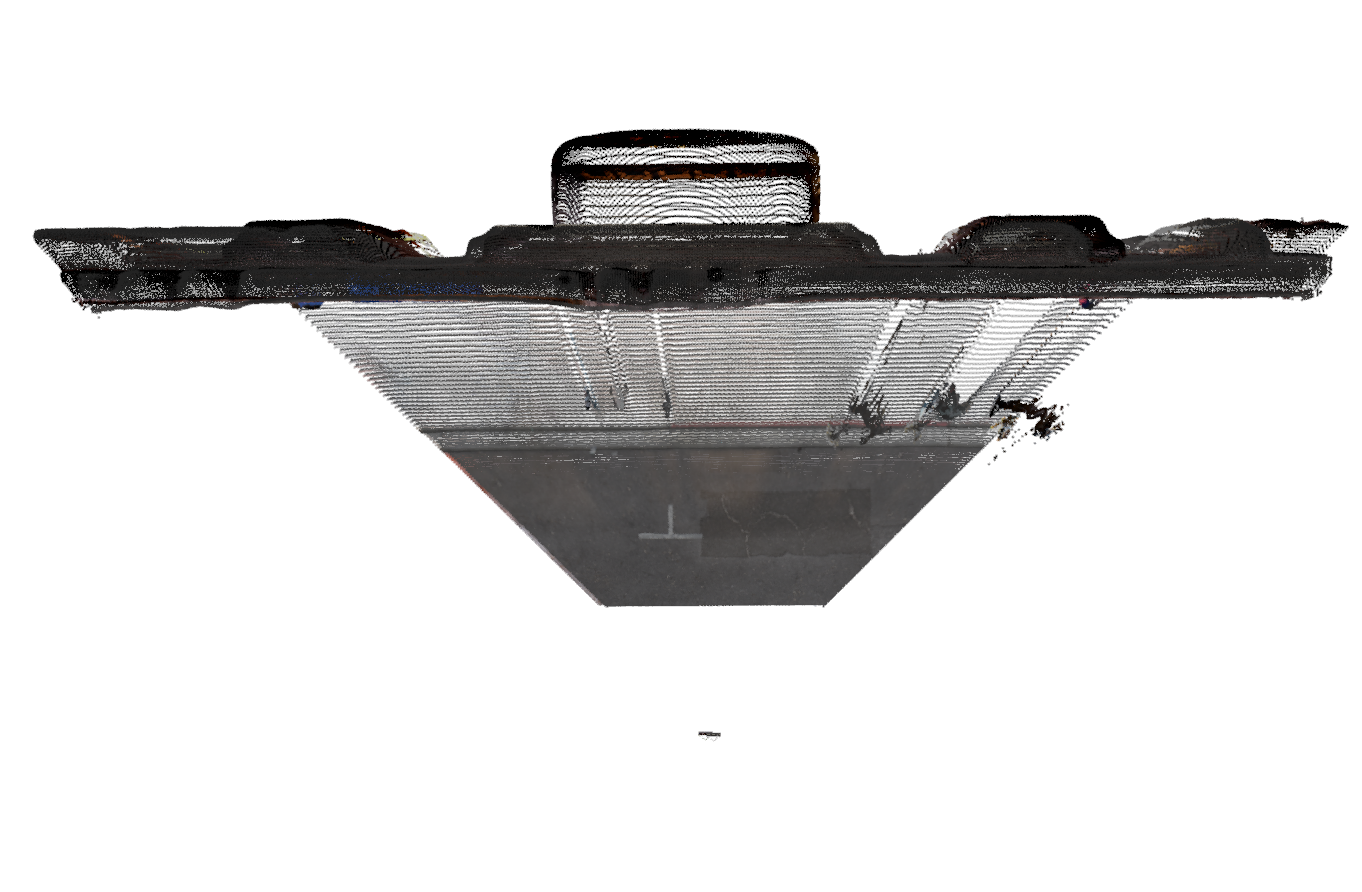}} \\
        \parbox[c]{0.45\linewidth}{\vspace{1pt}\centering\includegraphics[width=\linewidth]{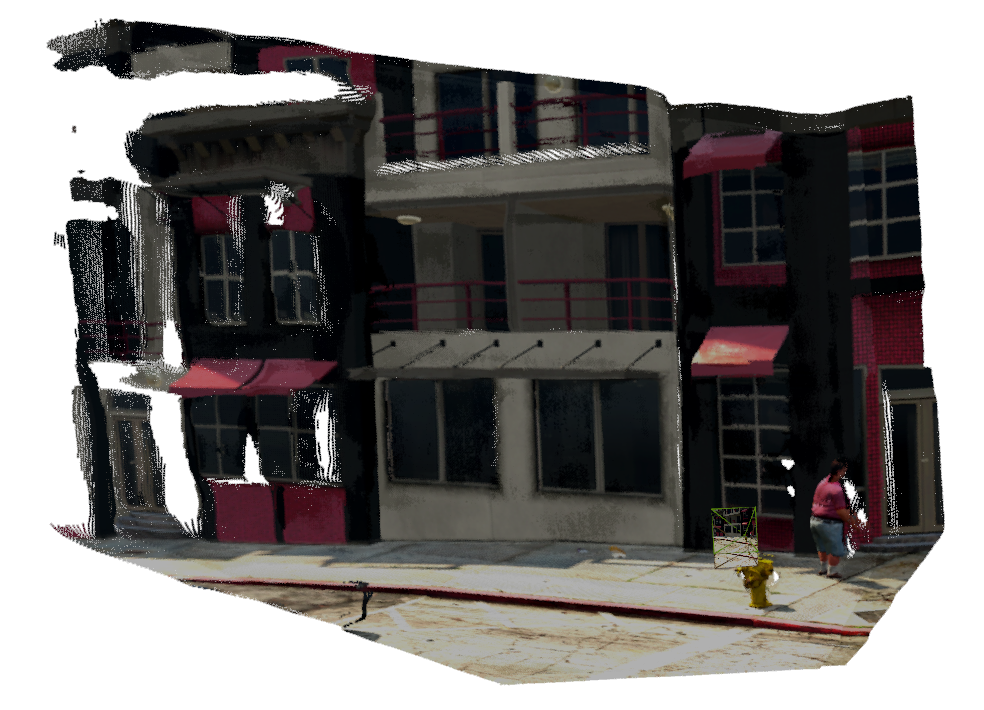}} &
        \parbox[c]{0.45\linewidth}{\vspace{3pt}\centering\includegraphics[width=\linewidth]{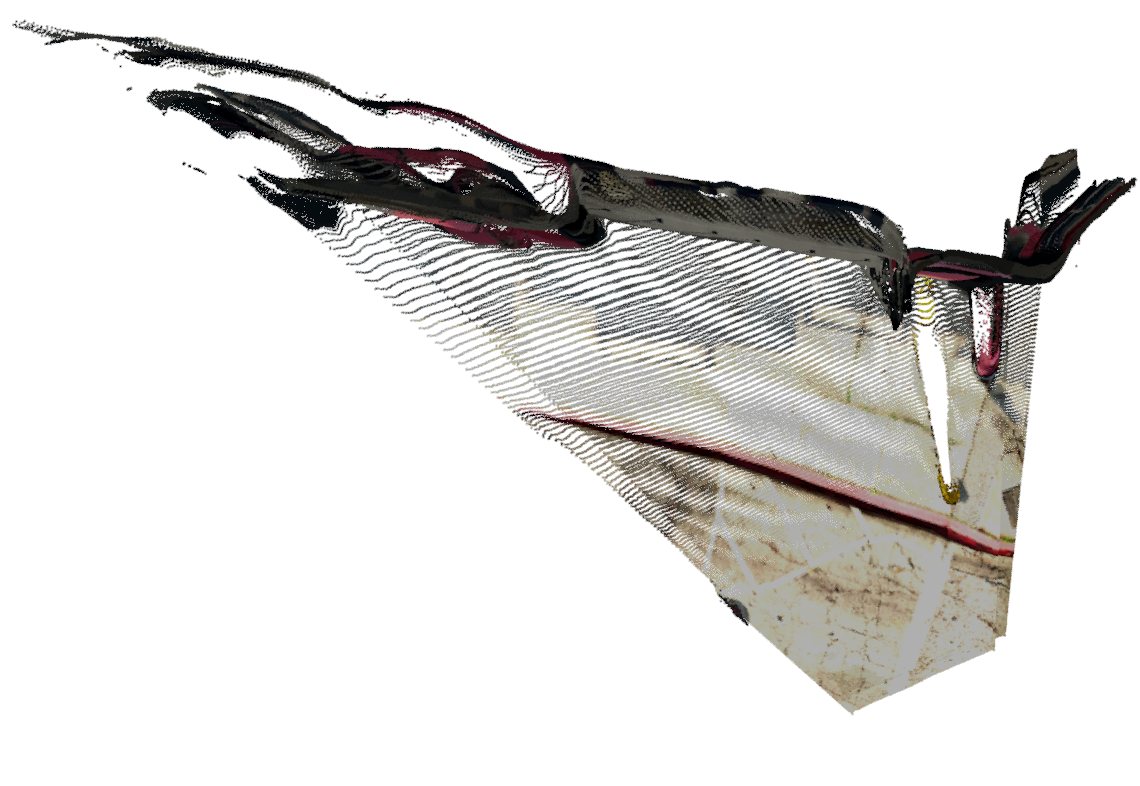}}

    \end{tabular}
    \vspace{-3pt}
    \caption{\textbf{Qualitative results of two-view 3D reconstruction on synthetic datasets.} We visualize the reconstructed point clouds from two different viewpoints (View 1 and View 2). For visualization purposes, the point clouds are colored using the ground truth RGB images.}
    \label{fig:suppl_Twoview_syn}
    \vspace{-1.2em}
\end{figure*}